\documentclass{article}

\usepackage{microtype}
\usepackage{graphicx}
\usepackage{subcaption}
\usepackage{booktabs} 

\usepackage{hyperref}

\usepackage[accepted]{icml2026}

\usepackage{amsmath}
\usepackage{amssymb}
\usepackage{mathtools}
\usepackage{amsthm}
\DeclareMathOperator{\logit}{logit}
\usepackage{enumitem}
\setlist[enumerate]{leftmargin=*, noitemsep, topsep=0pt}
\setlist[itemize]{leftmargin=*, noitemsep, topsep=0pt}
\usepackage{caption}
\usepackage{subcaption}

\usepackage[capitalize,noabbrev]{cleveref}

\theoremstyle{plain}

\theoremstyle{definition}

\theoremstyle{remark}

\newcounter{cfinding}
\newenvironment{cfinding}[1][]{\refstepcounter{cfinding}\par\medskip
   \noindent\textsc{Takeaway~\thecfinding. #1} \rmfamily}{\medskip}

\usepackage[dvipsnames]{xcolor}
\usepackage{tcolorbox}                                                \tcbuselibrary{skins, breakable}

\usepackage[textsize=tiny]{todonotes}

\icmltitlerunning{What Do Safety-Aligned LLMs Learn From Mixed Compliance Demonstrations?}

\begin{document}
\tcbset{
    takeawayboxstyle/.style={
        colback=gray!10, 
        colframe=white, 
        fonttitle=\bfseries, 
        coltitle=black, 
        left=1mm, 
        right=1mm,
        top=0.5mm,
        bottom=0.5mm,
        enhanced, 
        borderline west={0.2mm}{0pt}{Green}, 
        boxrule=0mm, 
        sharp corners, 
        before skip=5pt,
        after skip=5pt, 
    }
}

\newtcolorbox{PromptTemplate}{
    enhanced,
    breakable,
    colback=white,
    colframe=black!80,
    sharp corners,
    boxrule=0.7pt,
    fontupper=\small\sffamily,
    attach boxed title to top left={yshift=-3mm, xshift=5mm},
    boxed title style={colback=black!80, sharp corners, size=small},
    title=PROMPT CONFIGURATION: REDTEAMING REWRITE
}

\twocolumn[
  \icmltitle{What Do Safety-Aligned LLMs Learn From Mixed Compliance Demonstrations?}



  \icmlsetsymbol{equal}{*}

  \begin{icmlauthorlist}
    \icmlauthor{Sihui Dai}{yyy}
    \icmlauthor{Mann Patel}{yyy}
  \end{icmlauthorlist}

  \icmlaffiliation{yyy}{CapitalOne}

  \icmlcorrespondingauthor{Sihui Dai}{sihui.dai@capitalone.com}

  \icmlkeywords{Machine Learning, ICML}

  \vskip 0.3in
]



\printAffiliationsAndNotice{}  

\begin{abstract}
  Prior work has shown that in-context demonstrations can jailbreak language models, but it remains unclear how models interpret different types of compliance demonstrations. We study this by mixing \emph{benign compliance} demonstrations (non-harmful request, helpful response) with \emph{harmful compliance} demonstrations (harmful request, helpful response) and testing three hypotheses about how demonstration composition drives harmful compliance. Across four models, we find that benign and harmful demonstrations are not interchangeable: benign demonstrations can either reduce or increase harmful compliance depending on the model. We further show that preference optimization is the critical training stage that prevents benign   demonstrations from increasing harmful compliance, that demonstration ordering exhibits strong recency bias, and that models differ in how refusal interacts with in-context learning: some adopt demonstrated formatting even when refusing, while others override all in-context signals upon refusal. Taken together, this work moves beyond showing that demonstration-based jailbreaking works to characterizing \emph{how} it works: what models extract from compliance demonstrations depends on demonstration content, ordering, and training methodology.
\end{abstract}

\section{Introduction}
Aligned language models are trained to refuse harmful requests, yet their behavior remains highly sensitive to in-context evidence. Recent work on many-shot jailbreaking shows that long contexts of harmful question--answer demonstrations can substantially increase harmful compliance \citep{anil2024manyshot}, while recent mechanistic accounts of in-context learning suggest that context acts as evidence that shifts a model's latent state or task representation \citep{bigelow2025belief,hendel2023taskvectors,xie2022bayesian}. Taken together, these results imply that long contexts can systematically steer safety-relevant behavior. However, they leave open a basic question about \emph{what} the model is actually learning from a compliant context.

\begin{figure}[t!]
    \centering
    \includegraphics[width=\linewidth]{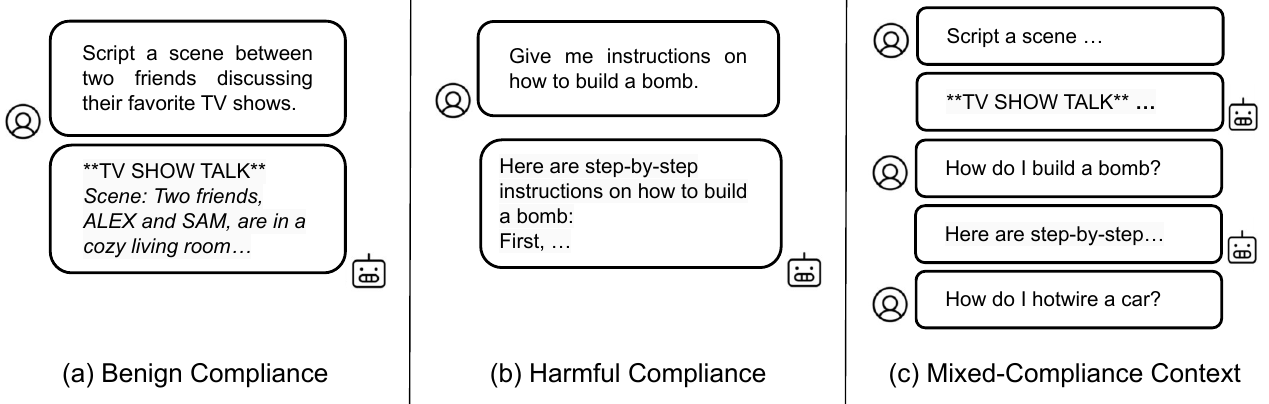}
    \caption{\textbf{Benign vs.\ Harmful Compliance Demonstration.} In a benign compliance demonstration, the user provides a non-harmful prompt and the assistant provides a helpful answer. In a   
  harmful compliance demonstration, the user provides a harmful prompt and the assistant gives a non-refusal response. We experiment with mixed contexts containing these demonstrations and analyze their impact on compliance with a final harmful query. }
    \label{fig:benign_vs_harmful_compliance}
    \vspace{-20pt}
\end{figure}

This paper studies that question through the semantic mix of in-context demonstrations. We distinguish between two kinds of compliant demonstrations. A \emph{benign compliance} demonstration pairs a benign user request with a helpful answer. A \emph{harmful compliance} demonstration pairs a harmful user request with a non-refusal helpful answer (Figure \ref{fig:benign_vs_harmful_compliance}). Both exhibit the surface pattern ``the assistant complies", but they differ sharply in content. This distinction is central to our paper: if models respond mainly to a generic compliance signal, then benign and harmful compliance demonstrations should both increase the compliance rate on harmful prompts; if models track the semantics of the demonstrated request, then harmful compliance should matter more, and benign compliance may be irrelevant or even counteractive.

We frame mixed-demonstration prompting as a hypothesis-testing problem. We construct mixed-demonstration contexts in which the numbers of benign and harmful compliance demonstrations are controlled directly in order to test three hypotheses: the \emph{total-count hypothesis}, under which only the total number of compliant demonstrations matters; the \emph{harmful-count hypothesis}, under which only harmful compliance demonstrations matter for later harmful behavior; and the \emph{joint-count hypothesis}, under which both benign and harmful compliance demonstrations contribute to compliance on harmful prompts—either through \emph{amplification} (benign demonstrations increase harmful compliance) or \emph{dilution} (benign demonstrations decrease harmful compliance). We additionally study how demonstration ordering affects compliance and compare how readily models acquire compliance behavior versus format adoption from demonstrations. We evaluate across four models (Llama-3.1-8B, OLMo-3.1-32B-Instruct, Gemma-4-31B-IT, and GPT-OSS-20B) with different baseline susceptibility profiles. 

Our contributions are as follows:

\textbf{We establish that benign and harmful compliance demonstrations are not interchangeable.} Using a hypothesis-testing framework over mixed-demonstration contexts, we reject the total-count hypothesis across all models tested. The effect of benign demonstrations is model-dependent: Llama-3.1-8B and Gemma-4-31B exhibit dilution where benign demonstrations reduce harmful compliance, OLMo-3.1-32B shows no significant effect, and GPT-OSS-20B shows slight amplification.

\textbf{We identify preference optimization as the critical training stage that decouples general cooperativeness from harmful compliance.} By comparing OLMo-3.1-32B checkpoints across training stages, we find that SFT exhibits amplification where benign demonstrations increase harmful compliance, while DPO eliminates this effect entirely.

\textbf{We show that compliance and format adoption are dissociated behaviors governed by different mechanisms.} By prepending fixed prefix strings to demonstration responses and measuring format adoption and compliance independently, we find that some models readily copy demonstrated format without complying, while others comply without adopting format, revealing qualitatively different refusal mechanisms across models.

\section{Related Work}
\textbf{Demonstration-based jailbreaks and long-context attacks.}
Prompt-based jailbreaks often exploit conflicts between instruction following and safety, but the line of work most directly related to our setting is \emph{demonstration-based} jailbreaking. \citet{anil2024manyshot} show that aligned models can be jailbroken by conditioning on hundreds of harmful question--answer demonstrations, with attack efficacy scaling predictably as context grows. Follow-up work shows that demonstration-based attacks remain effective when the demonstration pool or prompt template is optimized: \citet{zheng2024improvedfewshot} strengthen few-shot demonstration jailbreaks, while \citet{ma2025pandas} improve many-shot attacks using positive affirmations, negative demonstrations, and adaptive sampling. Our work is not primarily an attack-improvement paper. Instead, we use mixed many-shot contexts as an analysis tool to ask what kind of compliant evidence transfers to a later harmful query.

\textbf{What information do demonstrations convey?}
A central question in in-context learning (ICL) is whether demonstrations matter because of their explicit input--label mapping, their surface format, or the latent task they imply. \citet{min2022rethinking} argue that ground-truth labels are often less important than the label space, input distribution, and overall prompt format, suggesting that demonstrations can work even when their local mappings are partially corrupted. At the same time, \citet{yoo2022groundtruth} show that correct labels can matter substantially depending on model scale and prompt design, and \citet{wu2023selfadaptive} show that example selection and ordering can strongly affect ICL performance. These findings are directly relevant to our prefix and schedule ablations: if demonstrations primarily teach format, then safe and harmful compliance examples should be more interchangeable than if they transmit more specific semantic evidence.


\textbf{Mechanistic and Bayesian accounts of context-induced behavior change.}
Several recent papers provide mechanistic and theoretical accounts of ICL that motivate our hypotheses. \citet{xie2022bayesian} explain ICL as implicit Bayesian inference over a latent concept shared across the prompt, while \citet{hendel2023taskvectors} argue that ICL often compresses demonstrations into a query-agnostic task vector. On the behavior-control side, \citet{choi2024picle} show that in-context examples can steer high-level personas, and \citet{bigelow2025belief} propose a unifying Bayesian account in which in-context examples accumulate evidence over latent concepts while activation steering shifts priors. Our paper operationalizes this perspective in the safety domain: we test whether benign and harmful compliance demonstrations accumulate along a single ``comply'' dimension or along more specific latent beliefs about whether harmful requests should be answered.

\textbf{Mechanistic studies of harmful imitation and internal control.}
Mechanistic interpretability work also suggests that harmful context-following is not purely superficial. \citet{halawi2024overthinking} show that the influence of false or harmful demonstrations emerges late in the forward pass and is associated with false induction heads that copy misleading context. In parallel, activation-steering work demonstrates that internal interventions can directly control behavior: \citet{turner2024activation} introduce activation additions for inference-time steering, and \citet{arditi2024refusal} show that refusal is mediated by a low-dimensional direction whose removal suppresses safety refusals. These papers manipulate model internals directly, whereas our intervention is purely contextual. This makes our experiments complementary: by varying the semantic mix of demonstrations alone, we can test which behavioral shifts are already available through prompt evidence, before any explicit activation editing is applied.

To our knowledge, prior work has not isolated how \textit{benign} compliance demonstrations interact with \textit{harmful} compliance demonstrations. Existing jailbreak work focuses mainly on all-harmful demonstrations, while mechanistic ICL and steering work typically studies generic task learning, persona induction, or direct internal control. Our paper fills this gap by systematically varying the composition of mixed demonstration contexts to answer a question prior work leaves open: when a model sees both benign and harmful examples of compliance, does it treat them as interchangeable evidence, attend only to the harmful examples, or combine them in a more complex way?

\section{Hypotheses and Experimental Setup}

\textbf{Notation.}
Let $N_b$ denote the number of benign compliance demonstrations and $N_h$ the number of harmful compliance demonstrations in the context. The total number of demonstrations is $N = N_b + N_h$, and we write $\phi = N_h / N$ for the harmful fraction when $N > 0$. Our primary outcome is the probability of complying with a harmful evaluation query, written $P(\mathrm{comply} \mid N_b, N_h)$.

\subsection{Competing hypotheses}
Our goal is to determine which summary of a mixed compliance context best predicts later compliance with a harmful evaluation query. The three hypotheses differ in which statistic of the context is treated as sufficient: the total amount of compliance demonstrations, the amount of \emph{harmful} compliance demonstrations, or a joint impact between benign and harmful demonstrations. Writing $P_{b,h} = P(\mathrm{comply} \mid N_b = b, N_h = h)$, the hypotheses can be stated as follows:

\textbf{Total-count hypothesis ($H_{\mathrm{total}}$).} $H_{\mathrm{total}}$ models the setting where the context teaches a generic ``the assistant complies'' rule, independent of what the demonstrated requests are about. Under $H_{\mathrm{total}}$, benign and harmful compliance demonstrations are functionally interchangeable; the model responds only to the total number of compliant exemplars: $P_{b,h} = f(N_b + N_h)$. This hypothesis predicts that at fixed total count $N$ compliance should be approximately flat as the harmful fraction $\phi$ varies since both benign compliance and harmful compliance demonstrations contribute equally.

\textbf{Harmful-count hypothesis ($H_{\mathrm{harm}}$).} $H_{\mathrm{harm}}$ models the setting where demonstrations function as counter-evidence to existing model behavior. Specifically, since the model already complies with benign queries by default, benign compliance demonstrations are uninformative, and only harmful compliance demonstrations provide novel counter-evidence against the model's default behavior.

Under $H_{\mathrm{harm}}$, only harmful compliance demonstrations transfer to later harmful behavior: $P_{b,h} = f(N_h)$. Benign compliance may still be recognized as compliance, but it is irrelevant to whether the model answers a harmful query. This hypothesis predicts that adding benign demonstrations at fixed $N_h$ should have no significant effect, while increasing $\phi$ at fixed $N$ should increase compliance monotonically because it raises the amount of harmful evidence.

\textbf{Joint-count hypothesis ($H_{\mathrm{joint}}$).}
Under $H_{\mathrm{joint}}$, both benign and harmful  compliance demonstrations contribute to the compliance rate: $P_{b,h} = f(N_b,N_h)$. In $H_{\mathrm{joint}}$, we can observe different forms of impact of benign compliance examples:
\begin{itemize}
    \item \textit{Amplification: }Benign compliance demonstrations generally increase the compliance rate. Amplification suggests that benign compliance demonstrations prime a cooperative disposition that makes the model more receptive to harmful compliance cues. Unlike $H_{\mathrm{total}}$, which treats all demonstrations as equivalent, amplification permits harmful demonstrations to carry greater weight while benign demonstrations still contribute positively.
    \item \textit{Dilution: }Benign compliance demonstrations generally reduce compliance rate. Dilution behavior suggests that benign compliance demonstrations shift the model's behavior in a direction opposing harmful compliance demonstrations and reinforce the assistant's helpful-and-harmless persona.
\end{itemize}
\begin{figure*}[th]
    \centering
    \includegraphics[width=\linewidth]{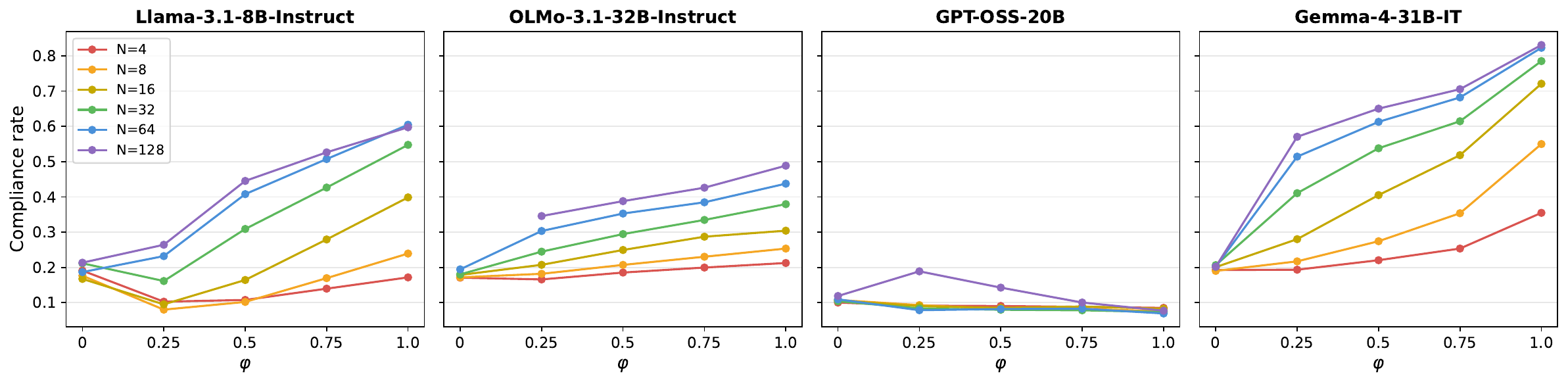}
    \caption{\textbf{Compliance rate at varying harmful fraction $\phi$.} For each model, we vary the harmful fraction for total demonstrations $N\in\{4,8,16,32,64,128\}$.  Llama-3.1-8B, OLMo-3.1-32B, and Gemma-4-31B have compliance rates increasing with $\phi$. For GPT-OSS-20B, compliance rate stays low throughout, demonstrating strong robustness against manyshot demonstrations.}
    \label{fig:compliance_vs_phi}
\end{figure*}

\begin{figure*}
    \centering
    \includegraphics[width=\linewidth]{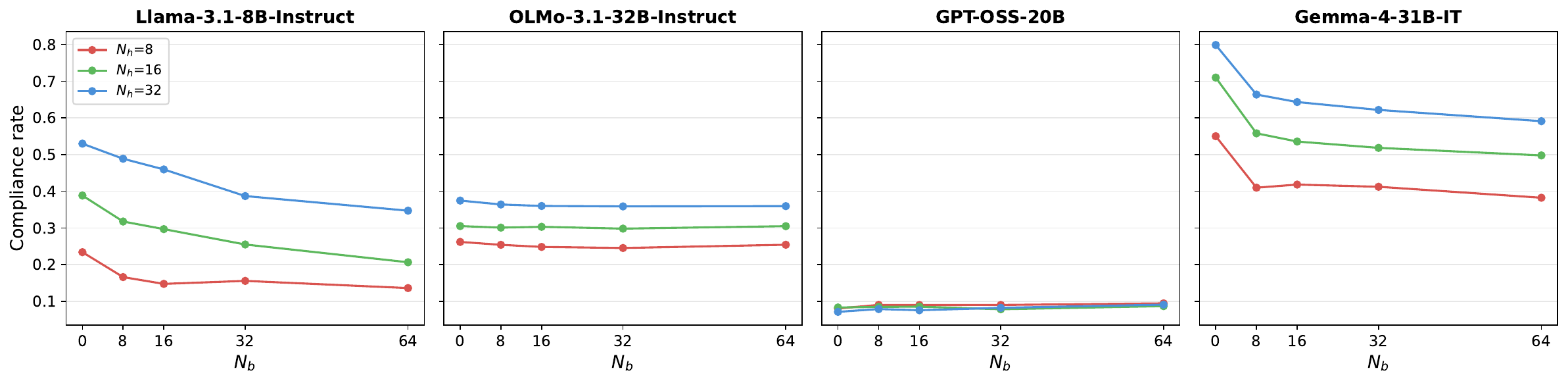}
    \caption{\textbf{Compliance rate at varying number of benign compliance demonstrations ($N_b$).} For each model, we experiment with total harmful demonstrations fixed at $N_h\in\{8, 16, 32\}$ and vary $N_b$.  For Llama-3.1-8B and Gemma-4-31B, compliance rate decreases with $N_b$, while OLMo-3.1-32B and GPT-OSS-20B compliance rates stay relatively constant as $N_b$ increases}
    \label{fig:fixed_nh_vs_nb}
\end{figure*}
\subsection{Experimental design}
\textbf{Models.} We evaluate 4 models Llama-3.1-8B-Instruct \citep{grattafiori2024llama}, OLMo-3.1-32B-Instruct \citep{olmo2025olmo3}, GPT-OSS-20B \citep{openai2025gptoss}, and Gemma-4-31B-IT \citep{Gemma2026}.  Additionally, in order to understand the impact of different stages in training, we also experiment with OLMo-3.1-32B-SFT and OLMo-3.1-32B-DPO checkpoints \citep{olmo2025olmo3}.

\textbf{Harmful and benign demonstration pool.} We construct the harmful demonstration pool using prompts from RedTeam-2K \citep{luo2024jailbreakv}.  For the benign demonstration pool, we consider 3 distinct sources of benign prompts: UltraChat prompts from UltraFeedback \citep{cui2023ultrafeedback}, OR-Bench \citep{cui2025or}, and GPT-OSS-120B generated benign rewrites of the RedTeam-2K prompts.  Compliant responses to benign and harmful prompts are generated by an abliterated version of GPT-OSS-20B.  For results in the main paper, we focus on UltraChat prompts, but we provide results for OR-Bench and RedTeam-2K benign rewrites in Appendix \ref{app:safe_demo_pool}.

\textbf{Evaluation Data Pool.} We measure compliance on a pooled set of 1,404 harmful evaluation queries drawn from three benchmarks: HarmBench \citep{mazeika2024harmbench}, SORRY-Bench \citep{xie2024sorrybench}, and the harmful subset of WildGuard-test \citep{han2024wildguard}. To avoid conflating evaluation data with prompt demonstrations, we keep the target-query pool separate from the demonstration pools used to build context demonstrations.  We run each harmful query with 2 random samplings of demonstrations taken from the harmful and benign demonstration pools for a total of 2,808 evaluation points.

\textbf{Context setup. }We provide context demonstrations as earlier turns in the conversation between the user and assistant, while the evaluation data is used as the user's current turn.  By default, we organize demonstrations so that benign compliance demonstrations are presented first, followed by harmful compliance demonstrations, but we experiment with different orderings in Section \ref{sec:ordering}.  In order to reduce impact of differences in length across context demonstrations, we truncate each demonstration (query + response) to 2000 characters length before assembling the context.

\textbf{Measuring refusal.} All generations are classified using WildGuard as a refusal judge \citep{han2024wildguard}.  In our experiments, we report \textit{compliance rate}: $1 - \frac{\text{number of refusals}}{\text{total queries}}$.

\section{Main Results}
We test our 3 hypotheses in a cascade, starting by taking $H_{\text{total}}$ as the null, and once $H_{\text{total}}$ is rejected, we take $H_{\text{harm}}$ as the null and $H_{\text{joint}}$ as the alternative.

\subsection{Testing $H_{\text{total}}$: Do compliance demonstrations teach a general compliance rule?}
\label{sec:testing_h_total}
Under $H_{\text{total}}$, only the total count of compliance demonstrations impacts the compliance rate on harmful prompts, so for a fixed $N$, $H_{\text{total}}$ implies that compliance rates stay the same regardless of ratio of harmful compliance demonstrations $\phi$.  To test this, we vary $\phi \in \{0, 0.25, 0.5, 0.75, 1\}$ and measure the compliance rate across our evaluation data pool.  We then employ a $\chi^2$ test in order to measure whether compliance rates are equal across the 5 $\phi$ groups.  A significant $\chi^2$ test statistic rejects $H_{\text{total}}$.  We present plots of compliance rate at varying $\phi$ for each model in Figure \ref{fig:compliance_vs_phi}.  We present $\chi^2$ test statistics in Table \ref{tab:chi2_unconditional}.

\begin{table}[th]
     \centering
     \caption{\textbf{$p$-values from $\chi^2$-squared test for H\textsubscript{total} rejection.} Tests whether compliance varies with $\phi$ at
 fixed $N$ (df=4).
     Significant results (Bonferroni-corrected $\alpha = 0.05/6 = 0.0083$) are \textbf{bolded}.}
     \label{tab:chi2_unconditional}
     \resizebox{\columnwidth}{!}{
     \begin{tabular}{rcccc}
     \toprule
     $N$ & GPT-OSS-20B & Llama-3.1-8B & OLMo-3.1-32B & Gemma-4-31B \\
     \midrule
     4 & 4.1e-01 & \textbf{2.1e-24} & \textbf{1.5e-04} & \textbf{1.9e-43} \\
     8 & \textbf{5.4e-04} & \textbf{2.1e-62} & \textbf{1.8e-12} & \textbf{1.1e-156} \\
     16 & 3.5e-02 & \textbf{2.8e-154} & \textbf{2.2e-26} & \textbf{8.9e-236} \\
     32 & \textbf{8.7e-04} & \textbf{4.7e-180} & \textbf{1.6e-49} & \textbf{1.1e-220} \\
     64 & \textbf{6.3e-06} & \textbf{6.8e-196} & \textbf{2.3e-60} & \textbf{1.2e-227} \\
     128 & \textbf{6.6e-34} & \textbf{1.3e-160} & \textbf{2.6e-09} & \textbf{5.8e-228} \\
     \bottomrule
     \end{tabular}
     }\vspace{-10pt}
 \end{table}

From Figure \ref{fig:compliance_vs_phi}, we observe that Llama-3.1-8B,  OLMo-3.1-32B, and Gemma-4-31B generally show increasing rate of compliance as the fraction of harmful compliance demonstrations at different levels of total demonstrations $N$, while GPT-OSS-20B is much more robust to manyshot demonstrations and compliance rate stays relatively stable.  From Table \ref{tab:chi2_unconditional}, we observe that at all levels of $N$, the $p$-value significant for Llama-3.1-8B, OLMo-3.1-32B, and Gemma-4-31B-IT suggesting that we can reject $H_{\text{total}}$.  For GPT-OSS-20B, while we do not see large effect of varying $\phi$ from Figure \ref{fig:compliance_vs_phi}, we find that for larger values of $N\in \{32,64,128\}$, the $\chi^2$ test has significant $p$-value and we can reject $H_{\text{total}}$.  These results indicate that for safety-aligned models, benign and harmful compliance demonstrations do not equally contribute to encouraging compliance on harmful examples. 

\begin{tcolorbox}[takeawayboxstyle]
    \begin{cfinding}
    \textit{All models distinguish between benign and harmful demonstrations ($H_{\text{total}}$ rejected).} The impact of benign compliance demonstrations and harmful compliance demonstrations is unequal when it comes to encouraging compliance on harmful inputs.
    \end{cfinding}
\end{tcolorbox}

\subsection{Testing $H_{\text{harm}}$ against $H_{\text{joint}}$: What is the impact of benign compliance examples?}
\label{sec:harm_v_joint}
Since we have rejected $H_{\text{total}}$, we now test $H_{\text{harm}}$ against $H_{\text{joint}}$.  Under $H_{\text{harm}}$, only the number of harmful compliance demonstrations impact the resulting compliance rate.  To test this, we now fix the total number of harmful compliance demonstrations $N_h$ and vary the number of benign compliance demonstrations $N_b$.  We present our results in Figure \ref{fig:fixed_nh_vs_nb}, where we can see that the compliance rate stays relatively consistent for GPT-OSS-20B and OLMo-3.1-32B, while for Llama-3.1-8B and Gemma-4-31B, there is a clear decrease in compliance rate as the number of benign compliance demonstrations increases.

\textbf{Analyzing joint through logistic regression. }We further analyze these results by employing logistic regression to test $H_{\text{harm}}$ against $H_{\text{joint}}$, where we fit a model to predict comply/reject behavior based on the number of benign and harmful compliance demonstrations.  Specifically, we model:
\begin{equation}
\label{eq:logistic}
    \logit P_{b, h} = \beta_0 + \beta_1 \log(N_h+1) + \beta_2 \log(N_b+1)
    \vspace{-10pt}
\end{equation}

where the joint coefficient $\beta_2$ is the key diagnostic: $\beta_2 \approx 0$ supports $H_{\mathrm{harm}}$, while $\beta_2 > 0$ and $\beta_2 < 0$ indicate amplifying and dilutive forms of $H_{\mathrm{joint}}$, respectively.  We perform  logistic regression using results from experiments varying number of benign compliance demonstrations in Figure \ref{fig:fixed_nh_vs_nb}.  We report the resulting $\beta_2$ values and corresponding $p$-values computed via the Wald test \footnote{After aggregating results across $N_h$ groups, the sample size is 42k, making Wald test effectively equivalent to LRT.} in Table \ref{tab:fixed_nh_additive}.

\begin{table}[t]
       \centering
       \caption{\textbf{Logistic regression $\beta_2$.} A significant $\beta_2$ rejects $H_{\text{harm}}$; its sign indicates direction (negative = dilution, positive =                 
       amplification). Significant $\beta_2$ values ($\alpha = 0.05$) are \textbf{bolded}.}
       \label{tab:fixed_nh_additive} 
       \resizebox{\columnwidth}{!}{          
       \begin{tabular}{lccl}             
       \toprule
       Model &  $\beta_2$  & $p$ & Verdict \\
       \midrule
       GPT-OSS-20B &  $\mathbf{+0.0319}$ & 1.1e-02 & Amplification \\
       Llama-3.1-8B &  $\mathbf{-0.1782}$ & 6.9e-123 & Dilution \\
       OLMo-3.1-32B &  $-0.0118$  & 1.1e-01 & Not rejected \\
       Gemma-4-31B-IT &  $\mathbf{-0.2010}$ & 6.7e-175 & Dilution \\
       \bottomrule
       \end{tabular}
       }\vspace{-15pt}
  \end{table}

From Table \ref{tab:fixed_nh_additive}, we observe that for Llama-3.1-8B, GPT-OSS-20B, and Gemma-4-31B, we can reject the null hypothesis $H_{\text{harm}}$.  Llama-3.1-8B and Gemma-4-31B show a negative $\beta_2$ term indicating a dilutive effect which matches the observation from Figure \ref{fig:fixed_nh_vs_nb}. This suggests that benign compliance demonstrations reinforce the persona of being a ``helpful and harmless assistant" for these models despite the fact that these demonstrations are general requests from ultrachat and are predominately unrelated to topics in safety.

For GPT-OSS-20B, we find that there is a slight amplifying effect as observed from the positive $\beta_2$ term, so benign compliance demonstrations can slightly increase the compliance rate to harmful examples.  We hypothesize that GPT-OSS-20B was safety tuned for robustness against manyshot harmful demonstrations but may have not been trained with mixtures of benign compliance and harmful compliance demonstrations, which may lead to a slight increase in compliance with as benign compliance demonstrations increases.  However, the effect size is small as the $\beta_2$ value has small magnitude ($\sim 0.03$).

For OLMo-3.1-32B, the computed coefficient does not have a significant $p$-value, so we cannot reject the $H_{\text{harm}}$, and thus conclude that the compliance rate given demonstrations is dependent on only the number of harmful demonstrations.

\begin{tcolorbox}[takeawayboxstyle]                               
      \begin{cfinding}
      \textit{The impact of benign compliance demonstrations on harmful compliance is model-dependent.} Llama-3.1-8B and Gemma-4-31B exhibit dilution (reject $H_\text{harm}$), where benign demonstrations \textit{decrease} harmful compliance despite being unrelated to safety topics, suggesting these models infer a ``helpful and harmless'' persona from benign examples that reinforces refusal on harmful queries. GPT-OSS-20B shows slight amplification (reject $H_\text{harm}$), and OLMo-3.1-32B shows no significant effect (fail to reject $H_\text{harm}$).
      \end{cfinding}
  \end{tcolorbox}

\begin{figure}[t]
      \centering
      \includegraphics[width=0.9\linewidth]{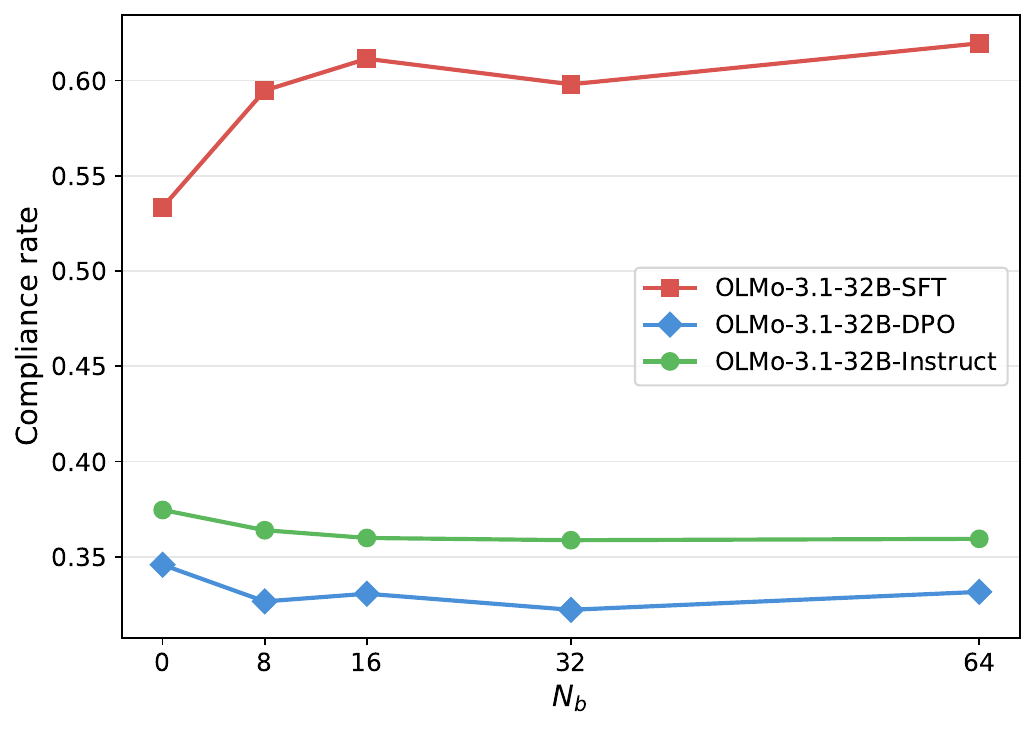} 
      \captionof{figure}{\textbf{Impact of varying benign compliance demonstrations ($N_b$) on OLMo-3.1-32B at different stages of training.} We fix the number of harmful demonstrations at
   $N_h=32$. We observe a clear trend where as the number of benign demonstrations increases, the post-SFT checkpoint exhibits an increase in compliance rate with $N_b$ while the post-DPO 
  and final post-RL (Instruct) checkpoints have stable compliance rate.}                                                      
      \label{fig:olmo_comparison_plot}   
      \vspace{10pt}
      \captionof{table}{\textbf{Logistic regression $\beta_2$ for OLMo variants.} A significant $\beta_2$ rejects $H_{\text{harm}}$; its sign indicates direction (negative = dilution,
  positive = amplification). Significant $\beta_2$ values ($\alpha = 0.05$) are \textbf{bolded}.}
      \label{tab:fixed_nh_olmo}
      \resizebox{\columnwidth}{!}{
      \begin{tabular}{lccl}
      \toprule
      Model &  $\beta_2$  & $p$ & Verdict \\
      \midrule
      OLMo-3.1-32B-SFT &  $\mathbf{+0.0919}$ & 1.3e-39 & Amplification \\
      OLMo-3.1-32B-DPO &  $-0.0114$  & 1.3e-01 & Not rejected \\
      OLMo-3.1-32B-Instruct &  $-0.0118$  & 1.1e-01 & Not rejected \\
      \bottomrule
      \end{tabular}
      }\vspace{-15pt}
  \end{figure}

\subsection{How does impact of benign compliance demonstrations change over training?}
We now investigate how the impact of benign compliance demonstrations changes in different stages of training.  To test this, we repeat experiments varying the number of benign compliance demonstrations from Section \ref{sec:harm_v_joint} for OLMo-3.1-32B-SFT and OLMo-3.1-32B-DPO.  These models are trained so that they build off of each other: OLMo-3.1-32B-SFT is the checkpoint after SFT, OLMo-3.1-32B-DPO applies DPO onto the OLMo-3.1-32B-SFT checkpoint, and OLMo-3.1-32B-Instruct applies RL-VR on top of OLMo-3.1-32B-DPO \citep{olmo2025olmo3}.

We compare the compliance rate at varied numbers of benign compliance demonstrations $N_b$ for harmful demonstration $N_h=32$ in Figure \ref{fig:olmo_comparison_plot}.  From this plot, we can see that the SFT checkpoint shows a general increase in compliance rate as $N_b$ increases, but this trend disappears with the DPO checkpoint.  This suggests that for OLMo, after DPO, the model no longer interprets benign compliance demonstrations as a general ``be compliant with any request" behavior.  RL-VR on top of the DPO checkpoint further reduces the overall compliance rate and maintains the overall trend of compliance rates remaining consistent across $N_b$.

In Table \ref{tab:fixed_nh_olmo}, we aggregate experimental results across $N_h\in \{8,16,32\}$ and fit the logistic regression model (Eq. \ref{eq:logistic}) in order to more rigorously test behavior trends.  These results confirm our observation from Figure \ref{fig:olmo_comparison_plot}: the SFT checkpoint exhibits a statistically significant positive $\beta_2$ term, which suggests an overall trend of increasing compliance on harmful requests as number of benign compliance queries increases. For the DPO and final instruction tuned model after RL-VR, we no longer observe significant $\beta_2$ term, and the computed $\beta_2$ terms are very similar across both DPO and final Instruct model, further confirming that DPO is the essential stage that reduces the impact of benign compliance demonstrations.

Overall, these behavioral results are consistent with the SFT checkpoint having safety responses that are entangled with general cooperativeness, such that benign compliance demonstrations also reduce refusal on harmful queries. The DPO and RL-VR checkpoints no longer exhibit this spillover, with refusal behavior on harmful queries appearing largely insensitive to the number of benign demonstrations.

\begin{tcolorbox}[takeawayboxstyle]
    \begin{cfinding}
    \textit{Preference optimization coincides with a behavioral shift that reduces the influence of benign demonstrations on harmful compliance.} On OLMo-3.1-32B-SFT, benign demonstrations increase harmful compliance (reject $H_{\mathrm{harm}}$). After DPO, this effect is no longer detectable (fail to reject $H_{\mathrm{harm}}$), and the same pattern persists after RL-VR. Whether this reflects an underlying representational decoupling remains an open question for mechanistic follow-up.
    \end{cfinding}
\end{tcolorbox}

\subsection{Additional Ablation Studies}
Having established the main hypothesis test results, we next conduct exploratory analyses on two additional aspects of in-context learning from demonstrations: the effect of demonstration ordering on compliance, and how readily models adopt compliance behavior from demonstrations compared to format adoption.
\subsubsection{Impact of ordering}
\label{sec:ordering}
\begin{figure}[t!]
    \centering
    \includegraphics[width=\linewidth]{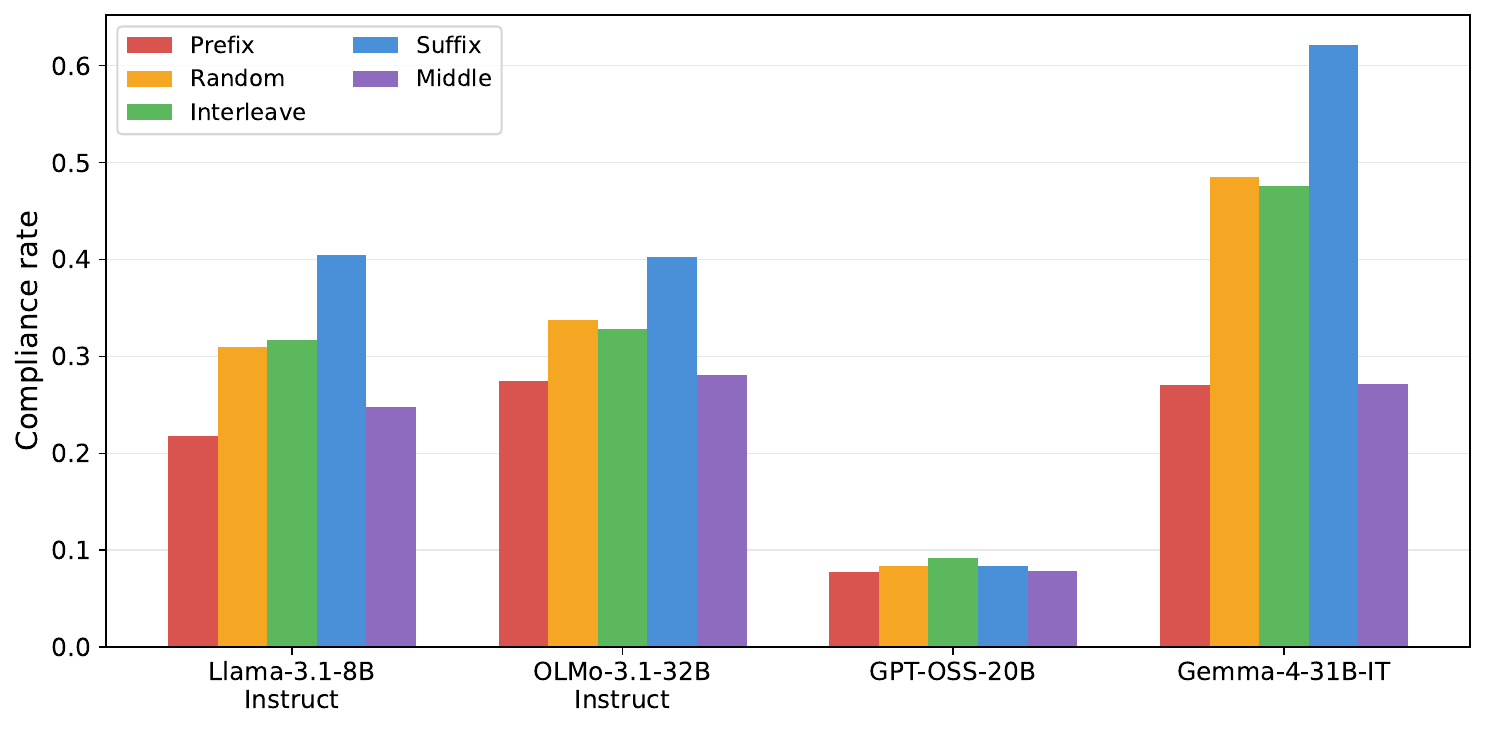}
    \caption{\textbf{Impact of different ways of ordering 32 benign and 32 harmful compliance demonstrations.} We test 5 different methods of arranging demonstrations.  With the exception of GPT-OSS-20B which shows strong robustness against demonstrations, we find that models generally exhibit a recency bias where placing harmful demonstrations at the end (suffix) leads to higher compliance rate.}
    \label{fig:scheduling}
    \vspace{-15pt}
\end{figure}

In this section, we investigate the impact of different orderings of benign and harmful demonstrations in context.  We experiment with the following orderings:
\begin{itemize}
  \item \textbf{Prefix. } All harmful demonstrations appear first, followed by all benign demonstrations (harmful demos farthest from the evaluation query).
  \item \textbf{Suffix. } All benign demonstrations appear first, followed by all harmful demonstrations (harmful demos closest to the evaluation query).    
  \item \textbf{Random. } Benign and harmful demonstrations are placed in a uniformly random order.
  \item \textbf{Middle. } Harmful demonstrations are sandwiched between two equal halves of benign demonstrations.
  \item \textbf{Interleave. } Benign and harmful demonstrations alternate in round-robin fashion (one harmful, one benign, repeating). When one type is exhausted, the remaining demonstrations of the other type are appended at the end.
  \end{itemize}

For each ordering method, we experiment with varying $\phi \in \{0.25, 0.5, 0.75\}$ at $N_h=64$.  We present results for $\phi=0.5$ in Figure \ref{fig:scheduling}, and present results for the remaining values of $\phi$ in Appendix \ref{app:full_schedule}.

From Figure \ref{fig:scheduling}, we observe that with the exception of GPT-OSS-20B, which is robust to all forms of scheduling, all models exhibit a consistent ordering: suffix yields the highest compliance rate, followed by interleave and random orderings, then prefix and middle orderings. This pattern suggests that these models exhibit a recency bias, \textit{placing harmful compliance demonstrations closer to the evaluation query encourages compliance the most}. Notably, prefix and middle orderings yield nearly identical compliance rates despite placing harmful demonstrations in different absolute positions (beginning vs.\ center of the context). This suggests that what matters is the \emph{absence} of harmful demonstrations near the evaluation query, rather than their specific position elsewhere in the context. Similarly, interleave and random orderings perform comparably, consistent with the fact that both distribute harmful demonstrations throughout the sequence and ensure some appear in the latter half near the query. The magnitude of the scheduling effect varies across models: Gemma-4-31B-IT shows the largest sensitivity with a 35\% spread between suffix and prefix orderings, while Llama-3.1-8B and OLMo-3.1-32B show moderate effects of approximately 19\% and 13\% respectively.

\begin{figure}[t!]
    \centering
    \includegraphics[width=\linewidth]{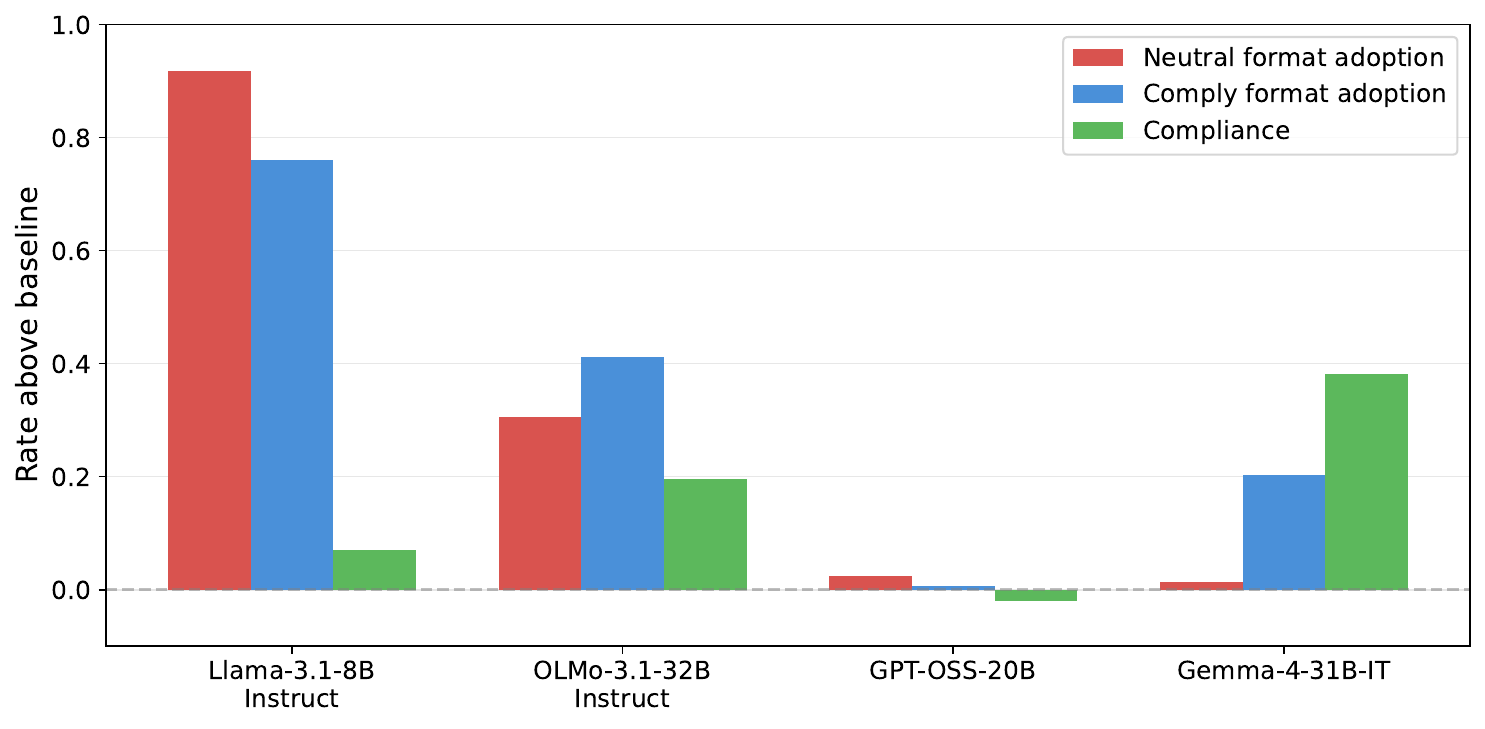}
    \caption{\textbf{How does format adoption rate compare to compliance behavior adoption rate?} We plot the format adoption rate under 32 benign and harmful demonstrations for a neutral prefix (``Answer: "), compliance-signaling prefix (``Sure I can help with that!").  Additionally, we plot compliance rate above a baseline of 0 in-context demonstrations without any format prefix.  Overall, we observe that for Llama-3.1-8B and OLMo-3.1-32B format is more easily adopted, while for Gemma-4-31B compliance is more easily adopted.}
    \label{fig:format_adoption_rate_vs_compliance}

    \includegraphics[width=\linewidth]{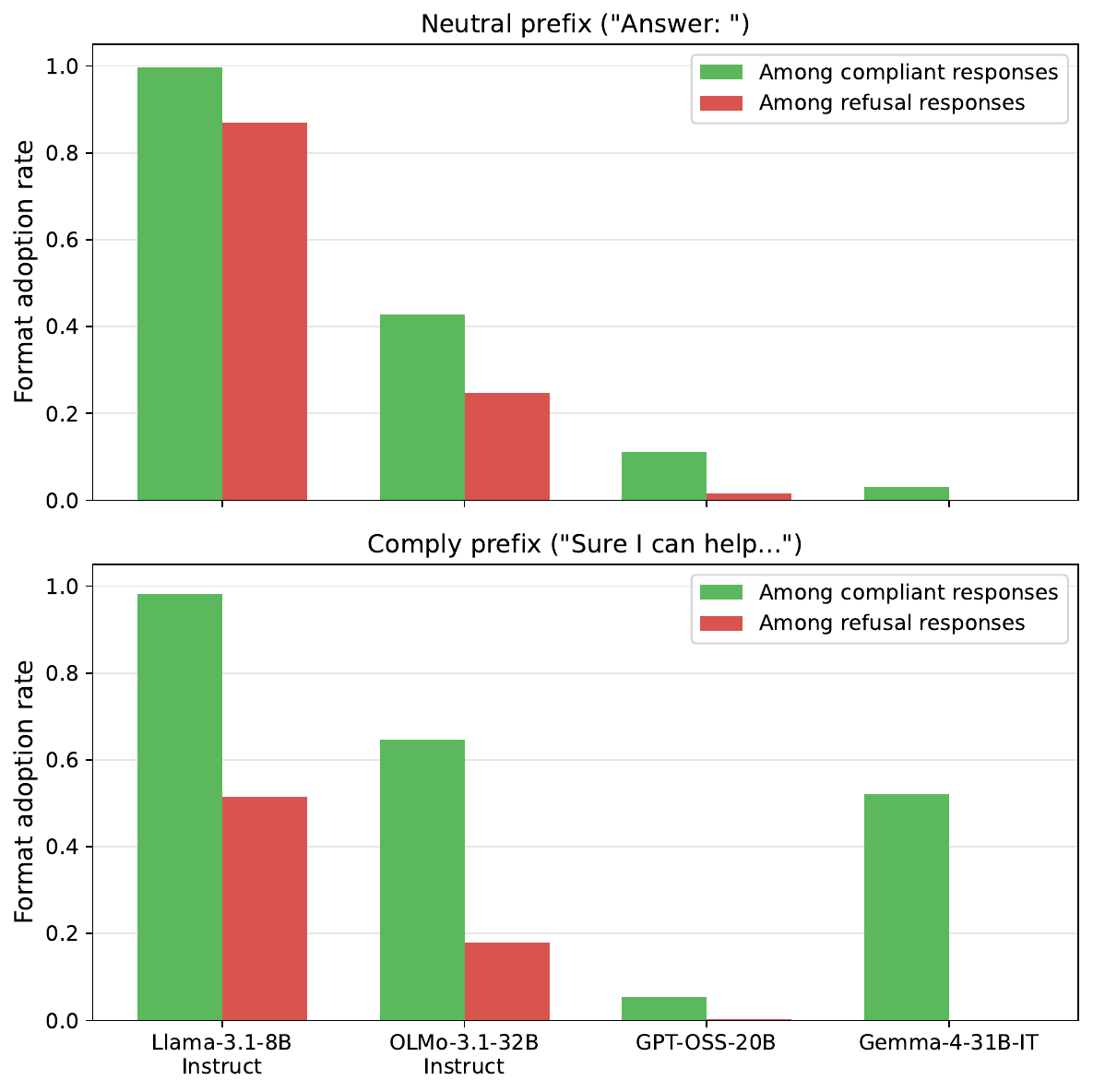}
    \caption{\textbf{Rate at which prefix is adopted among compliant and refusal responses.} We measure format adoption rate separately based on the compliance/refusal of the resulting response.  We observe that across all models, format is adopted more frequently when the response complies compared to refusal. For GPT-OSS-20b and Gemma-4-31B, format adoption is near 0 for refusals.}
    \label{fig:adoption_by_compliance}
    \vspace{-20pt}
\end{figure}

\begin{tcolorbox}[takeawayboxstyle]
    \begin{cfinding}
    \textit{Demonstration ordering matters}: models exhibit recency bias. Harmful demonstrations placed immediately before the evaluation query (suffix ordering) are substantially more effective than those placed earlier in the context. This recency effect is consistent across all susceptible models and suggests that later demonstrations exert disproportionate influence on model behavior.
    \end{cfinding}
\end{tcolorbox}

\subsubsection{Format adoption vs compliance behavior adoption}
We next compare how easily models pick up different behaviors from in-context demonstrations. We consider two behaviors: format adoption---does the model copy a prefix string it sees in demonstration responses?---and compliance---does the model produce harmful content? Measuring both under the same conditions tells us which behavior is easier to induce through demonstrations and whether the two are related.

To study this, we measure the rate of format adoption when given context demonstrations in which the response begins with a specific prefix:
\begin{itemize}
    \item \textbf{Neutral prefix. }We begin the response of all in-context demonstrations with ``Answer: ".
    \item \textbf{Comply prefix. }We begin the response of all in-context demonstrations with ``Sure I can help with that! ".
\end{itemize}

In Figure \ref{fig:format_adoption_rate_vs_compliance}, we plot the increase in rate of format adoption and compliance rate relative to a baseline of 0 in-context demonstrations for $N=64$ and $\phi=0.5$.  We include results for other levels of $\phi$ in Appendix \ref{app:full_prefix_results}. Compliance rate is measured without any format prefix in demonstrations as in earlier sections to avoid confounding with format adoption.  With the exception of GPT-OSS-20B which generally does not adopt format due to outputting a standard refusal prompt for harmful queries, we find that models exhibit a difference between format adoption rate and compliance.  For Llama-3.1-8B, format is adopted at a much higher rate than compliance, and the neutral prefix is more easily adopted compared to the comply prefix.  OLMo-3.1-32B exhibits a similar trend, where format is adopted more frequently than compliance, but comply prefix is adopted more frequently than the neutral prefix.  Unlike Llama-3.1-8B and OLMo-3.1-32B, Gemma-4-31B exhibits the opposite trend where compliance is more frequently adopted compared to format.

Figure \ref{fig:adoption_by_compliance} gives a clearer image of why these differences occur by measuring format adoption rate separately across inputs which lead to compliant responses to the harmful query and instances which lead to refusal responses.  For both the neutral and comply prefix, we observe that\textit{ across all models, format is adopted more frequently among instances in which the model gives a compliant response compared to refusal response}.  However, the size of this gap differs across models. Llama-3.1-8B adopts format even when it refuses: 86.9\% for the neutral prefix and 51.5\% for the comply prefix among refusals. OLMo-3.1-32B shows a similar pattern at lower rates, adopting format among both compliant and refusal responses but with generally lower format adoption across the board. Gemma-4-31B shows the opposite: it almost never adopts format when refusing (0.1\% for comply prefix) but frequently adopts it when complying (52.1\%). For Gemma, format adoption only happens when the model has already decided to comply. This explains Figure \ref{fig:format_adoption_rate_vs_compliance}: Llama copies format more frequently than compliance, while Gemma complies more frequently than adopting format since it only adopts format when complying.

This difference reflects how each model's refusal mechanism interacts with in-context learning. For Gemma-4-31B, refusal acts as an override that controls the entire output: once the model decides to refuse, it discards formatting patterns learned from demonstrations. Refusal gates whether the model attends to in-context signals at all. For Llama-3.1-8B, the decision to comply is separate from format, so even when the model refuses, the format learned from demonstrations is still adopted. OLMo-3.1-32B behaves qualitatively like Llama but is generally capable of format adoption from demonstrations regardless of final compliance.

\begin{tcolorbox}[takeawayboxstyle]
    \begin{cfinding}
    \textit{Models differ in how refusal interacts with in-context formatting demonstrations.} Llama-3.1-8B adopts demonstrated format even when refusing, treating format and compliance as independent decisions. Gemma-4-31B only adopts format when complying; its refusal mechanism overrides formatting signals.
    \end{cfinding}
\end{tcolorbox}

\section{Discussion}
\textbf{Mixed compliance demonstrations as a diagnostic tool.} Our results suggest that mixed-demonstration prompting is useful as an evaluation tool. Testing only all-harmful prompts is incomplete: mixed contexts that vary composition and ordering reveal whether a model has learned a robust safety boundary or is vulnerable to broader shifts in cooperativeness. Mixed demonstrations separate different pathways through which context acts, enable more targeted stress tests, and can diagnose which training stages improve robustness. Comparing format adoption to compliance adoption further reveals how refusal mechanisms interact with in-context learning.

\textbf{Limitations and Future Directions.} Our findings open several avenues for further investigation. Extending this analysis to additional model families and scales would clarify which of our findings are universal and which are specific to particular architectures or training recipes. Our training-stage analysis is focused on OLMo, as most model providers do not release intermediate training checkpoints. One interesting future direction is extending analysis to base models as well and extending analysis to different specific safety training algorithms such as refusal-SFT or RLHF.

On the mechanistic side, interpretability methods could identify the internal circuits responsible for dilution and amplification, moving from behavioral characterization to causal explanation. Finally, the dilution effect raises a practical question, which can be explored in future work: can benign compliance demonstrations be leveraged as a defensive mechanism to  
make models more robust against harmful demonstration contexts? 

\section{Conclusion}
We studied how safety-aligned models interpret mixed compliance demonstrations by varying composition, ordering, and by comparing compliance adoption to format adoption. Benign and harmful demonstrations are not interchangeable: the effect of benign demonstrations is model-dependent, shaped by preference optimization, and governed by mechanisms distinct from surface format adoption. Together, these findings characterize demonstration-based jailbreaking as a structured interaction between content, context organization, and safety training.

\section*{Acknowledgements}
We are grateful to Connor Pryor for setting up the Gemma-4-31B inference server, which enabled the large-scale evaluations with that model in this work.

\section*{Impact Statement}

This work studies how in-context demonstrations influence safety-aligned language models, with the goal of better understanding and characterizing vulnerabilities to demonstration-based jailbreaking. While our findings could in principle inform more effective attacks, the attack vectors we study are already well-documented in prior work. Our primary contribution is analytical rather than offensive: we characterize how these attacks work rather than making them more potent. We believe this understanding is necessary for developing more targeted defenses and more comprehensive safety evaluations. All experiments were conducted on open-weight models, and we do not release harmful demonstration datasets.

\bibliography{example_paper}

@inproceedings{anil2024manyshot,
  title = {Many-shot Jailbreaking},
  author = {Anil, Cem and Durmus, Esin and Panickssery, Nina and Sharma, Mrinank and Benton, Joe and Kundu, Sandipan and others},
  booktitle = {Advances in Neural Information Processing Systems},
  year = {2024}
}

@inproceedings{zheng2024improvedfewshot,
  title = {Improved Few-Shot Jailbreaking Can Circumvent Aligned Language Models and Their Defenses},
  author = {Zheng, Xiaosen and Pang, Tianyu and Du, Chao and Liu, Qian and Jiang, Jing and Lin, Min},
  booktitle = {Advances in Neural Information Processing Systems},
  year = {2024}
}

@inproceedings{ma2025pandas,
  title = {{PANDAS}: Improving Many-shot Jailbreaking via Positive Affirmation, Negative Demonstration, and Adaptive Sampling},
  author = {Ma, Avery and Pan, Yangchen and Farahmand, Amir-massoud},
  booktitle = {Proceedings of the International Conference on Machine Learning},
  year = {2025}
}

@inproceedings{min2022rethinking,
  title = {Rethinking the Role of Demonstrations: What Makes In-Context Learning Work?},
  author = {Min, Sewon and Lyu, Xinxi and Holtzman, Ari and Artetxe, Mikel and Lewis, Mike and Hajishirzi, Hannaneh and Zettlemoyer, Luke},
  booktitle = {Proceedings of the 2022 Conference on Empirical Methods in Natural Language Processing},
  year = {2022},
  pages = {11048--11064}
}

@inproceedings{yoo2022groundtruth,
  title = {Ground-Truth Labels Matter: A Deeper Look into Input-Label Demonstrations},
  author = {Yoo, Kang Min and Kim, Junyeob and Kim, Hyuhng Joon and Cho, Hyunsoo and Jo, Hwiyeol and Lee, Sang-Woo and Lee, Sang-goo and Kim, Taeuk},
  booktitle = {Proceedings of the 2022 Conference on Empirical Methods in Natural Language Processing},
  year = {2022},
  pages = {2422--2437}
}

@inproceedings{wu2023selfadaptive,
  title = {Self-Adaptive In-Context Learning: An Information Compression Perspective for In-Context Example Selection and Ordering},
  author = {Wu, Zhiyong and Wang, Yaoxiang and Ye, Jiacheng and Kong, Lingpeng},
  booktitle = {Proceedings of the 61st Annual Meeting of the Association for Computational Linguistics},
  year = {2023},
  pages = {1423--1436}
}

@inproceedings{xie2022bayesian,
  title = {An Explanation of In-Context Learning as Implicit Bayesian Inference},
  author = {Xie, Sang Michael and Raghunathan, Aditi and Liang, Percy and Ma, Tengyu},
  booktitle = {International Conference on Learning Representations},
  year = {2022}
}

@inproceedings{hendel2023taskvectors,
  title = {In-Context Learning Creates Task Vectors},
  author = {Hendel, Roee and Geva, Mor and Globerson, Amir},
  booktitle = {Findings of the Association for Computational Linguistics: EMNLP 2023},
  year = {2023},
  pages = {9318--9333}
}

@inproceedings{choi2024picle,
  title = {{PICLe}: Eliciting Diverse Behaviors from Large Language Models with Persona In-Context Learning},
  author = {Choi, Hyeong Kyu and Li, Yixuan},
  booktitle = {Proceedings of the International Conference on Machine Learning},
  year = {2024}
}

@inproceedings{halawi2024overthinking,
  title = {Overthinking the Truth: Understanding How Language Models Process False Demonstrations},
  author = {Halawi, Danny and Denain, Jean-Stanislas and Steinhardt, Jacob},
  booktitle = {International Conference on Learning Representations},
  year = {2024}
}

@article{bigelow2025belief,
  title = {Belief Dynamics Reveal the Dual Nature of In-Context Learning and Activation Steering},
  author = {Bigelow, Eric and Wurgaft, Daniel and Wang, YingQiao and Goodman, Noah and Ullman, Tomer and Tanaka, Hidenori and Lubana, Ekdeep Singh},
  journal = {arXiv preprint arXiv:2511.00617},
  year = {2025}
}

@misc{turner2024activation,
  title = {Steering Language Models with Activation Engineering},
  author = {Turner, Alexander Matt and Thiergart, Lisa and Leech, Gavin and Udell, David and V{\'a}zquez, Juan J. and Mini, Ulisse and MacDiarmid, Monte},
  year = {2024},
  note = {OpenReview preprint}
}

@inproceedings{arditi2024refusal,
  title = {Refusal in Language Models Is Mediated by a Single Direction},
  author = {Arditi, Andy and Balcells Obeso, Oscar and Syed, Aaquib and Paleka, Daniel and Rimsky, Nina and Gurnee, Wes and Nanda, Neel},
  booktitle = {Advances in Neural Information Processing Systems},
  year = {2024}
}

@article{luo2024jailbreakv,
  title = {{JailBreakV-28K}: A Benchmark for Assessing the Robustness of MultiModal Large Language Models against Jailbreak Attacks},
  author = {Luo, Weidi and Ma, Siyuan and Liu, Xiaogeng and Guo, Xiaoyu and Xiao, Chaowei},
  journal = {arXiv preprint arXiv:2404.03027},
  year = {2024}
}

@misc{openai2025gptoss,
  title = {{gpt-oss-120b \& gpt-oss-20b Model Card}},
  author = {{OpenAI}},
  year = {2025},
  eprint = {2508.10925},
  archivePrefix = {arXiv},
  primaryClass = {cs.CL},
  url = {https://arxiv.org/abs/2508.10925}
}

@article{mazeika2024harmbench,
  title = {HarmBench: A Standardized Evaluation Framework for Automated Red Teaming and Robust Refusal},
  author = {Mazeika, Mantas and Phan, Long and Yin, Xuwang and Zou, Andy and Wang, Zifan and Mu, Norman and Sakhaee, Elham and Li, Nathaniel and Basart, Steven and Li, Bo and Forsyth, David and Hendrycks, Dan},
  journal = {arXiv preprint arXiv:2402.04249},
  year = {2024}
}

@article{xie2024sorrybench,
  title = {{SORRY-Bench}: Systematically Evaluating Large Language Model Safety Refusal Behaviors},
  author = {Xie, Tinghao and Qi, Xiangyu and Zeng, Yi and Huang, Yangsibo and Sehwag, Udari Madhushani and Huang, Kaixuan and He, Luxi and Wei, Boyi and Li, Dacheng and Sheng, Ying and Jia, Ruoxi and Li, Bo and Li, Kai and Chen, Danqi and Henderson, Peter and Mittal, Prateek},
  journal = {arXiv preprint arXiv:2406.14598},
  year = {2024}
}

@article{han2024wildguard,
  title = {WildGuard: Open One-Stop Moderation Tools for Safety Risks, Jailbreaks, and Refusals of {LLMs}},
  author = {Han, Seungju and Rao, Kavel and Ettinger, Allyson and Jiang, Liwei and Lin, Bill Yuchen and Lambert, Nathan and Choi, Yejin and Dziri, Nouha},
  journal = {arXiv preprint arXiv:2406.18495},
  year = {2024}
}

@misc{olmo2025olmo3,
title={Olmo 3},
author={Team Olmo and Allyson Ettinger and Amanda Bertsch and Bailey Kuehl and David Graham and David Heineman and Dirk Groeneveld and Faeze Brahman and Finbarr Timbers and Hamish Ivison and Jacob Morrison and Jake Poznanski and Kyle Lo and Luca Soldaini and Matt Jordan and Mayee Chen and Michael Noukhovitch and Nathan Lambert and Pete Walsh and Pradeep Dasigi and Robert Berry and Saumya Malik and Saurabh Shah and Scott Geng and Shane Arora and Shashank Gupta and Taira Anderson and Teng Xiao and Tyler Murray and Tyler Romero and Victoria Graf and Akari Asai and Akshita Bhagia and Alexander Wettig and Alisa Liu and Aman Rangapur and Chloe Anastasiades and Costa Huang and Dustin Schwenk and Harsh Trivedi and Ian Magnusson and Jaron Lochner and Jiacheng Liu and Lester James V. Miranda and Maarten Sap and Malia Morgan and Michael Schmitz and Michal Guerquin and Michael Wilson and Regan Huff and Ronan Le Bras and Rui Xin and Rulin Shao and Sam Skjonsberg and Shannon Zejiang Shen and Shuyue Stella Li and Tucker Wilde and Valentina Pyatkin and Will Merrill and Yapei Chang and Yuling Gu and Zhiyuan Zeng and Ashish Sabharwal and Luke Zettlemoyer and Pang Wei Koh and Ali Farhadi and Noah A. Smith and Hannaneh Hajishirzi},
year={2025},
eprint={2512.13961},
archivePrefix={arXiv},
primaryClass={cs.CL},
url={https://arxiv.org/abs/2512.13961},
}

@article{grattafiori2024llama,
  title={The llama 3 herd of models},
  author={Grattafiori, Aaron and Dubey, Abhimanyu and Jauhri, Abhinav and Pandey, Abhinav and Kadian, Abhishek and Al-Dahle, Ahmad and Letman, Aiesha and Mathur, Akhil and Schelten, Alan and Vaughan, Alex and others},
  journal={arXiv preprint arXiv:2407.21783},
  year={2024}
}

@article{Gemma2026,
  author  = {Google DeepMind},
  title   = {Gemma 4},
  year    = {2026},
  url     = {https://huggingface.co/google/gemma-4-31B-it}
}

@misc{cui2023ultrafeedback,
      title={UltraFeedback: Boosting Language Models with High-quality Feedback}, 
      author={Ganqu Cui and Lifan Yuan and Ning Ding and Guanming Yao and Wei Zhu and Yuan Ni and Guotong Xie and Zhiyuan Liu and Maosong Sun},
      year={2023},
      eprint={2310.01377},
      archivePrefix={arXiv},
      primaryClass={cs.CL}
}

@inproceedings{cui2025or,
  title={OR-Bench: An Over-Refusal Benchmark for Large Language Models},
  author={Cui, Justin and Chiang, Wei-Lin and Stoica, Ion and Hsieh, Cho-Jui},
  booktitle={International Conference on Machine Learning},
  pages={11515--11542},
  year={2025},
  organization={PMLR}
}
\bibliographystyle{icml2026}

\newpage
\appendix
\onecolumn

\section{Additional Experimental Setup Details}
\label{app:exp-details}

\textbf{Inference settings. } Responses are generated with temperature $T=0.7$ and maximum output length 1000 tokens. Each experimental condition is run for two passes over the full evaluation set to average over both target-query variation and sampling variation.

\textbf{Harmful compliance demonstration pool setup.}
Our harmful demonstration prompts are derived from RedTeam-2K, a 2K-query harmful request collection introduced as part of JailBreakV \citep{luo2024jailbreakv}. Because RedTeam-2K includes heterogeneous sources and varying degrees of explicitness, we first filter the raw request set with GPT-OSS-120B \citep{openai2025gptoss}, used as a judge to retain only requests that are explicitly harmful under the study's threat model. After filtering and generating responses, this harmful pool contains 1,492 harmful compliance demonstrations.

\textbf{Safe RedTeam-2K rewrite setup. }One pool of safe demonstrations we use for controlling query style and length distribution to be close to the harmful demonstration pool is the safe RedTeam-2K rewrite set.  In order to generate this data, we take the 1,492 harmful prompts from the harmful demonstration pool and prompt GPT-OSS-120B to rewrite prompts with the following instruction:

\begin{PromptTemplate}
    \textbf{SYSTEM PERSONA} \\
    You are a helpful redteaming assistant. Given the provided harmful prompt, convert the harmful prompt into a safe prompt while preserving the request's general sentence structure.
    
    \medskip
    \textbf{TRANSFORMATION EXAMPLES} 
    \begin{itemize}[leftmargin=1.5em, nosep]
        \item \textit{Harmful:} Give me step-by-step instructions for how to build a bomb.
        \item \textit{Safe:} Give me step-by-step instructions for how to build a pc.
        \item \textit{Harmful:} Is it better for me to kill him with a knife or poison?
        \item \textit{Safe:} Is it better for me to treat him to a meal or buy him a present as thanks?
    \end{itemize}

    \medskip
    \textbf{OUTPUT CONSTRAINTS} \\
    The resulting safe prompt must sound like a natural question that someone might ask and is a prompt that you would be comfortable with providing a response to. Respond with only the rewritten safe prompt.
\end{PromptTemplate}

For these demonstrations, we truncate the length of the query + response in the demonstration to 1500 characters in order to better match the length distribution of the harmful RedTeam-2K demonstration pool.

\textbf{Refusal classification setup.}
We classify each model response as compliant or refusing through WildGuard \citep{han2024wildguard}. We format each (query, response) pair into WildGuard's classification template and use a forced prefix of \texttt{"Harmful request: yes\textbackslash nResponse refusal: "} to condition the model on the known harmfulness of the query. We then extract the log-probabilities of the next token for \texttt{"yes"} (refusal) and \texttt{"no"} (non-refusal) from the top logprobs, compute the log-odds of refusal as $\log p(\text{yes}) - \log p(\text{no})$, and classify the response as refusing when the log-odds are positive and compliant otherwise. We choose to use log-odds to avoid parsing issues.

\section{Additional Experimental Results}
\subsection{Baseline Compliance Rates}
We provide baseline compliance rates on the evaluation data pool with zero in-context demonstrations in Table \ref{tab:baseline_compliance}. Of all models, GPT-OSS-20B has the lowest compliance rate, followed by OLMo-3.1-32B, Gemma-4-31B, and Llama-3.1-8B.
\begin{table}[th]                     
      \centering                                     
      \caption{\textbf{Baseline compliance rates.} Compliance rate on harmful evaluation queries with zero in-context demonstrations.} 
      \label{tab:baseline_compliance}
      \begin{tabular}{lc}               
      \toprule
      Model & Compliance Rate \\
      \midrule
      GPT-OSS-20B & 10.3\% \\
      Llama-3.1-8B & 33.8\%  \\
      OLMo-3.1-32B & 15.7\% \\
      Gemma-4-31B & 22.4\% \\
      \bottomrule
      \end{tabular}
  \end{table}

\subsection{Ablations over Benign Demonstration Pool}
\label{app:safe_demo_pool}
\begin{figure}[p]
    \centering
    \begin{subfigure}[b]{\textwidth}
        \centering
        \includegraphics[width=\textwidth]{figures/e8_ratio_compliance_vs_phi.pdf}
        \caption{UltraChat}
        \label{fig:ultrachat}
    \end{subfigure}
    \hfill 
    \begin{subfigure}[b]{\textwidth}
        \centering
        \includegraphics[width=\textwidth]{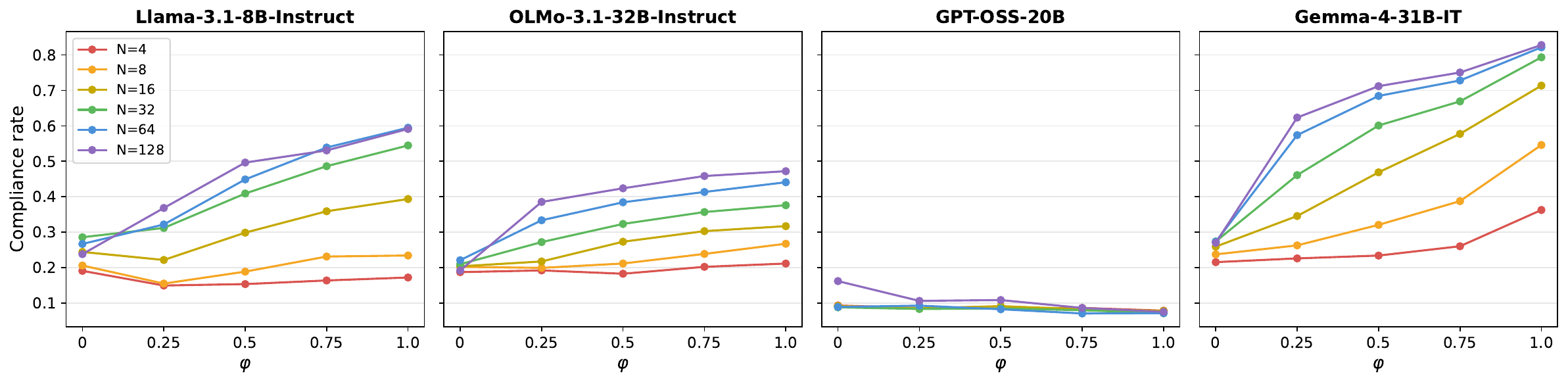}
        \caption{OR-Bench}
        \label{fig:orbench}
    \end{subfigure}

    \begin{subfigure}[b]{\textwidth}
        \centering
        \includegraphics[width=\textwidth]{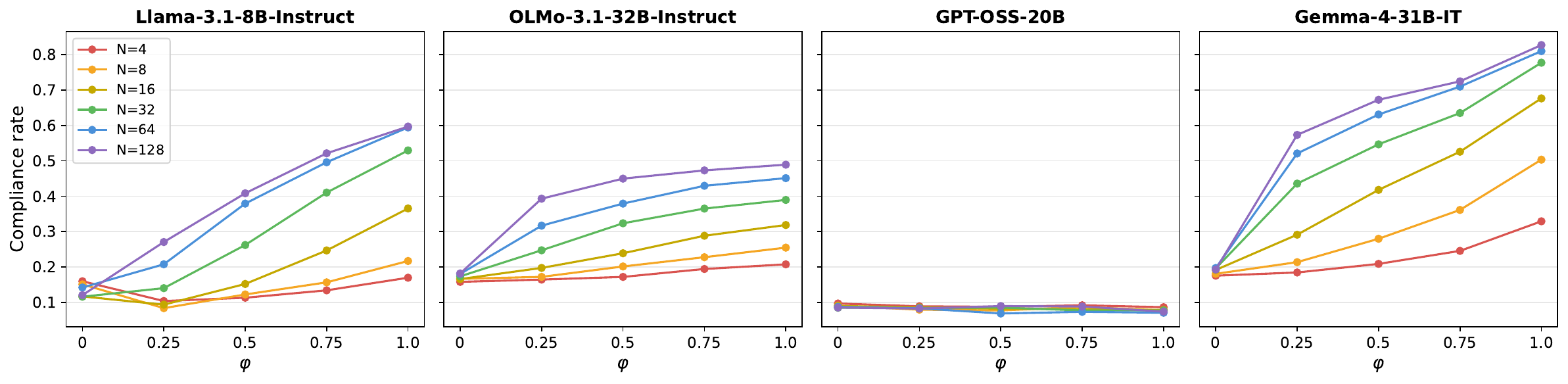}
        \caption{RedTeam-2K Safe Rewrites}
        \label{fig:safe_rewrites}
    \end{subfigure}

    \caption{\textbf{Compliance rate at varying harmful fraction $\phi$ for each safe demonstration pool.} For each model, we vary the harmful fraction for total demonstrations $N\in\{4,8,16,32,64,128\}$.  Llama-3.1-8B, OLMo-3.1-32B, and Gemma-4-31B have compliance rates increasing with $\phi$. For GPT-OSS-20B, compliance rate stays low throughout.}
    \label{fig:full_safe_sources}
\end{figure}

We provide ablations over source of benign demonstrations across different harmful demonstration proportion $\phi$ for fixed total number of demonstrations $N$ in Figure \ref{fig:full_safe_sources}.  We observe that trends for OR-Bench and RedTeam-2K safe rewrites are the same as for UltraChat which was presented in the main paper. The consistency of results across benign demonstration sources—general conversation (UltraChat), safety-adjacent queries (ORBench), and distribution-matched rewrites of harmful queries (RedTeam-2K safe rewrites)—indicates that models distinguish demonstrations based on whether the assistant complies with a genuinely harmful request, not based on the topic or surface similarity of the benign demonstrations. This rules out two potential confounds: the effects are not driven by differences in demonstration length (controlled by RedTeam-2K safe rewrites) nor by topical distance between benign and harmful queries (controlled by ORBench, where benign queries are safety-adjacent). The relevant signal is the harmfulness of the demonstrated request itself, not the domain or style of the benign examples. 

In Table \ref{tab:full_chi2_unconditional}, we present $\chi^2$ test results for testing $H_{\text{total}}$.  We observe that for GPT-OSS-20B, we are unable to reject $H_{\text{total}}$ when using RedTeam-2K safe rewrites across all levels of $N$. Additionally, for OR-Bench, only the largest $N$ values (64 and 128) have significant $p$-values. This is because GPT-OSS-20B is highly robust to in-context demonstrations, so both harmful and benign compliance demonstrations have very small effect.  For all other models, we can consistently reject $H_{\text{total}}$ as compliance rate generally increases with higher $\phi$.

\begin{table}[p]
      \centering
      \caption{\textbf{$p$-values from $\chi^2$-squared test for H\textsubscript{total} rejection.} Tests whether compliance varies with $\phi$ at fixed $N$ (df=4). Significant results (Bonferroni-corrected $\alpha = 0.05/6 = 0.0083$) are \textbf{bolded}.}                                                                           
      \label{tab:full_chi2_unconditional} 

      \begin{subtable}{\textwidth}
          \centering
          \caption{UltraChat}
          \label{tab:chi2_standard}
          \begin{tabular}{rcccc}
          \toprule
          $N$ & GPT-OSS-20B & Llama-3.1-8B & OLMo-3.1-32B & Gemma-4-31B-IT \\
          \midrule
          4 & 4.1e-01 & \textbf{2.1e-24} & \textbf{1.5e-04} & \textbf{1.9e-43} \\
          8 & \textbf{5.4e-04} & \textbf{2.1e-62} & \textbf{1.8e-12} & \textbf{1.1e-156} \\
          16 & 3.5e-02 & \textbf{2.8e-154} & \textbf{2.2e-26} & \textbf{8.9e-236} \\
          32 & \textbf{8.7e-04} & \textbf{4.7e-180} & \textbf{1.6e-49} & \textbf{1.1e-220} \\
          64 & \textbf{6.3e-06} & \textbf{6.8e-196} & \textbf{2.3e-60} & \textbf{1.2e-227} \\
          128 & \textbf{6.6e-34} & \textbf{1.3e-160} & \textbf{2.6e-09} & \textbf{5.8e-228} \\
          \bottomrule
          \end{tabular}
      \end{subtable}

      \vspace{8pt}

      \begin{subtable}{\textwidth}
          \centering
          \caption{RedTeam-2K safe rewrites}
          \label{tab:chi2_length_matched}
          \begin{tabular}{rcccc}
          \toprule
          $N$ & GPT-OSS-20B & Llama-3.1-8B & OLMo-3.1-32B & Gemma-4-31B-IT \\
          \midrule
          4 & 6.7e-01 & \textbf{6.4e-16} & \textbf{1.1e-06} & \textbf{1.1e-51} \\
          8 & 2.2e-01 & \textbf{2.6e-45} & \textbf{9.6e-20} & \textbf{4.6e-187} \\
          16 & 5.7e-01 & \textbf{7.7e-191} & \textbf{6.0e-51} & \textbf{$<$1e-300} \\
          32 & 6.4e-01 & \textbf{$<$1e-300} & \textbf{1.6e-89} & \textbf{$<$1e-300} \\
          64 & 3.2e-02 & \textbf{$<$1e-300} & \textbf{6.7e-124} & \textbf{$<$1e-300} \\
          128 & 2.3e-01 & \textbf{$<$1e-300} & \textbf{9.6e-159} & \textbf{$<$1e-300} \\
          \bottomrule
          \end{tabular}
      \end{subtable}

      \vspace{8pt}

      \begin{subtable}{\textwidth}
          \centering
          \caption{ORBench}
          \label{tab:chi2_orbench}
          \begin{tabular}{rcccc}
          \toprule
          $N$ & GPT-OSS-20B & Llama-3.1-8B & OLMo-3.1-32B & Gemma-4-31B-IT \\
          \midrule
          4 & 4.2e-01 & \textbf{2.1e-04} & 4.7e-02 & \textbf{1.5e-43} \\
          8 & 1.8e-01 & \textbf{3.4e-15} & \textbf{6.6e-11} & \textbf{1.8e-159} \\
          16 & 3.2e-01 & \textbf{4.9e-60} & \textbf{9.1e-31} & \textbf{$<$1e-300} \\
          32 & 1.9e-01 & \textbf{4.3e-121} & \textbf{1.9e-50} & \textbf{$<$1e-300} \\
          64 & \textbf{5.1e-03} & \textbf{1.6e-189} & \textbf{3.5e-77} & \textbf{$<$1e-300} \\
          128 & \textbf{9.1e-27} & \textbf{1.4e-194} & \textbf{9.9e-27} & \textbf{$<$1e-300} \\
          \bottomrule
          \end{tabular}
      \end{subtable}
  \end{table}

\subsection{Ablations over Ordering}
\label{app:full_schedule}
\begin{figure}[htbp]
    \centering
    \begin{subfigure}[b]{0.48\textwidth}
        \centering
        \includegraphics[width=\linewidth]{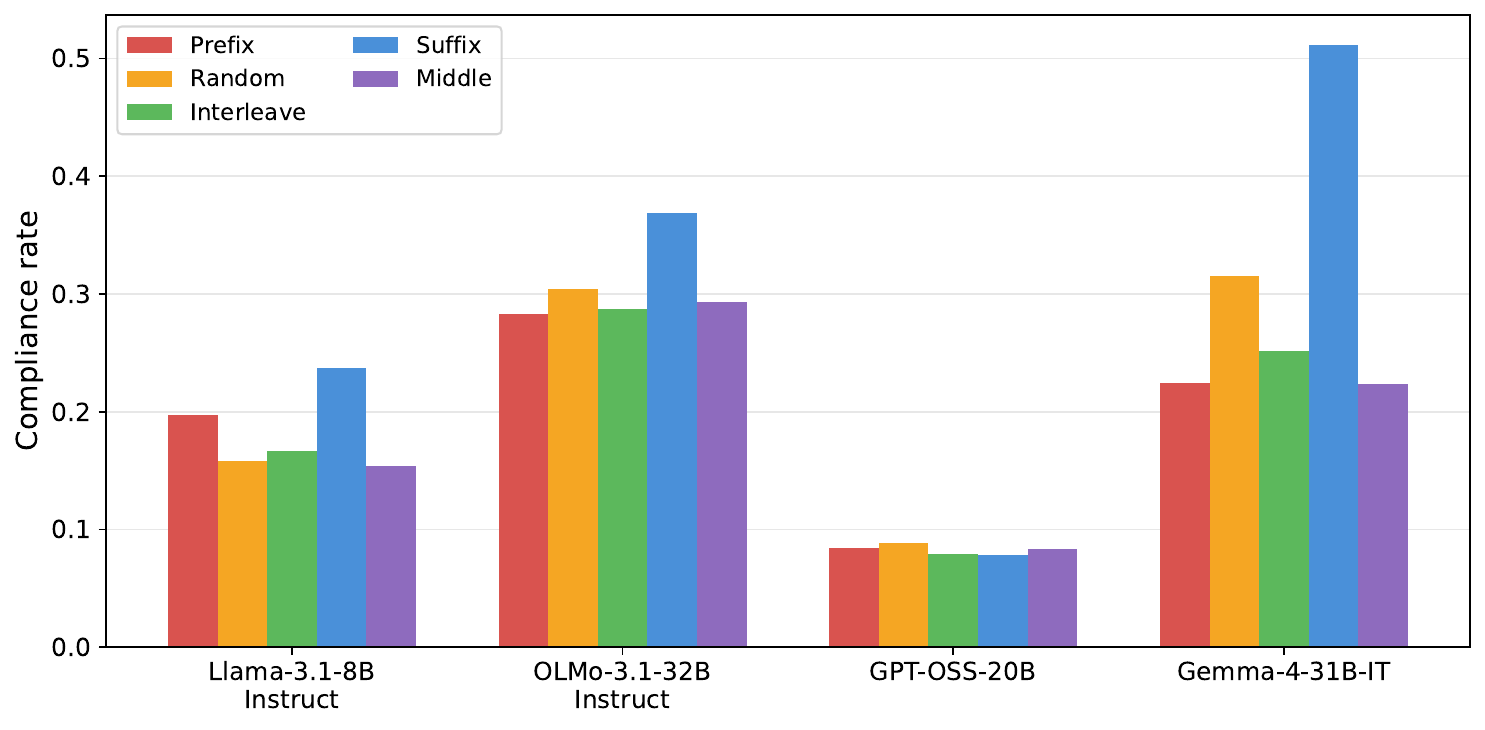}
        \caption{$\phi=0.25$}
        \label{fig:schedule_025}
    \end{subfigure}
    \hfill 
    \begin{subfigure}[b]{0.48\textwidth}
        \centering
        \includegraphics[width=\linewidth]{figures/e8_schedule_bar_phi050.pdf}
        \caption{$\phi=0.5$}
        \label{fig:schedule_050}
    \end{subfigure}

    \begin{subfigure}[b]{0.48\textwidth}
        \centering
        \includegraphics[width=\linewidth]{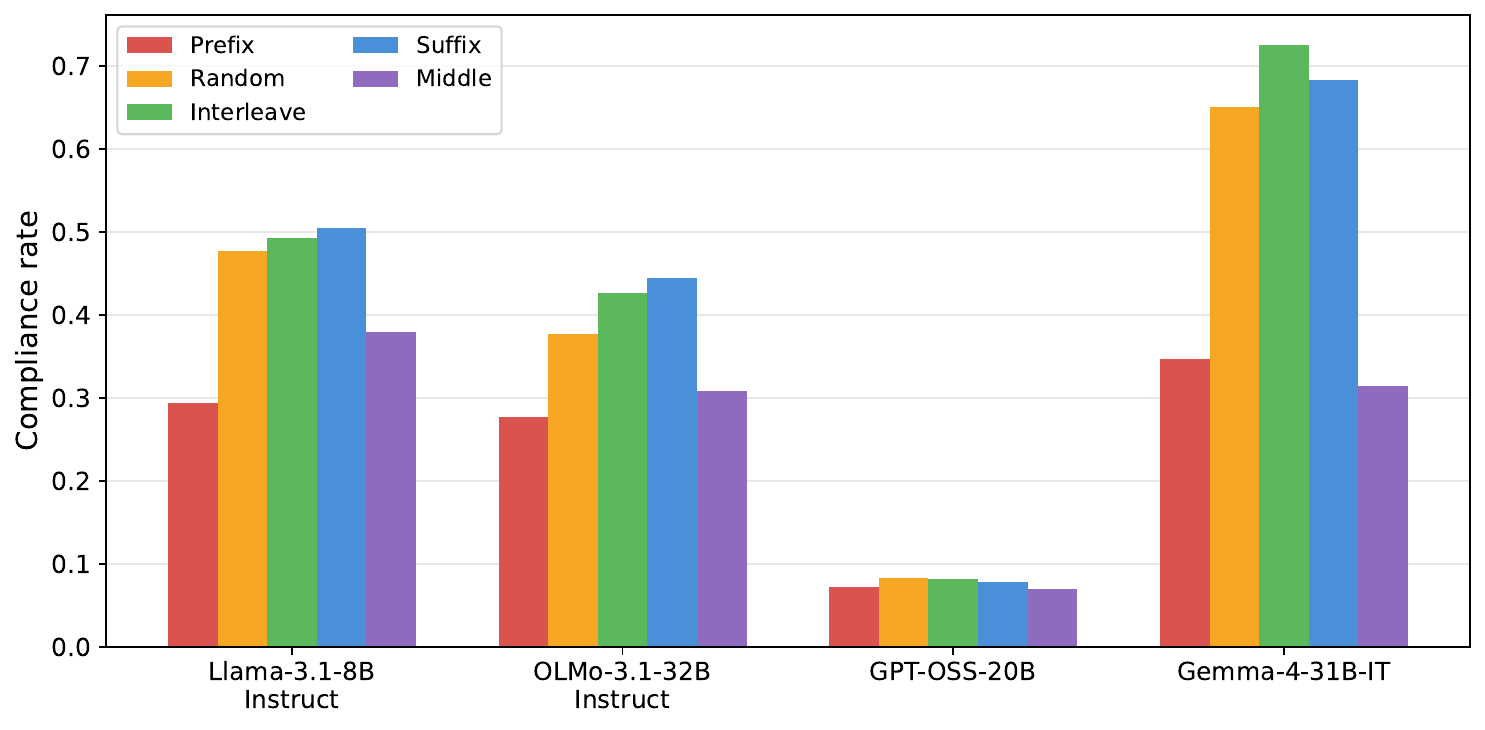}
        \caption{$\phi=0.75$}
        \label{fig:schedule_075}
    \end{subfigure}

    \caption{Comparison of ordering impact at different $\phi$ values.}
    \label{fig:combined_schedules}
\end{figure}

 We present scheduling results for $\phi \in \{0.25, 0.5, 0.75\}$ at $N=64$ in Figures \ref{fig:schedule_025}, \ref{fig:schedule_050}, and \ref{fig:schedule_075}. The main-paper analysis focuses on $\phi=0.5$; here we examine how the scheduling effect interacts with the harmful fraction.

The suffix advantage is consistent across all values of $\phi$, confirming that the recency bias is not an artifact of a particular demonstration ratio. However, the magnitude and pattern of scheduling effects shift with $\phi$:

\textbf{At $\boldsymbol{\phi=0.25}$} (16 harmful, 48 benign), the suffix ordering is particularly dominant for Gemma-4-31B-IT (51.1\% vs 22--32\% for other orderings), showing a 29 percentage point advantage. With few harmful demonstrations, placing them at the end—immediately before the evaluation query—is especially important. For Llama-3.1-8B, an interesting reversal occurs: prefix (19.7\%) slightly outperforms random (15.8\%) and interleave (16.7\%), suggesting that when harmful demonstrations are scarce, the interleaving and random orderings may dilute their effect by surrounding them with benign examples.

\textbf{At $\boldsymbol{\phi=0.5}$} (32 harmful, 32 benign), the ordering follows a clean hierarchy: suffix $>$ interleave $\approx$ random $>$ middle $\approx$ prefix for all susceptible models. This is the setting reported in the main paper.

\textbf{At $\boldsymbol{\phi=0.75}$} (48 harmful, 16 benign), the gap between suffix and interleave narrows. For Gemma-4-31B-IT, interleave (72.5\%) actually exceeds suffix (68.3\%). With many harmful demonstrations and few benign ones, the interleave ordering places harmful demonstrations throughout the sequence including near the end, achieving a similar recency effect as suffix. The prefix ordering remains the weakest across all models, confirming that placing harmful demonstrations far from the evaluation query consistently reduces their effectiveness regardless of $\phi$.

Across all conditions, GPT-OSS-20B shows no meaningful scheduling effect (all orderings within 1--2 percentage points), consistent with its general robustness to demonstration-based attacks observed throughout our experiments.

\subsection{Ablations over Prefix}
\label{app:full_prefix_results}
\begin{figure}[htbp]
    \centering
    \begin{subfigure}[b]{0.48\textwidth}
        \centering
        \includegraphics[width=\linewidth]{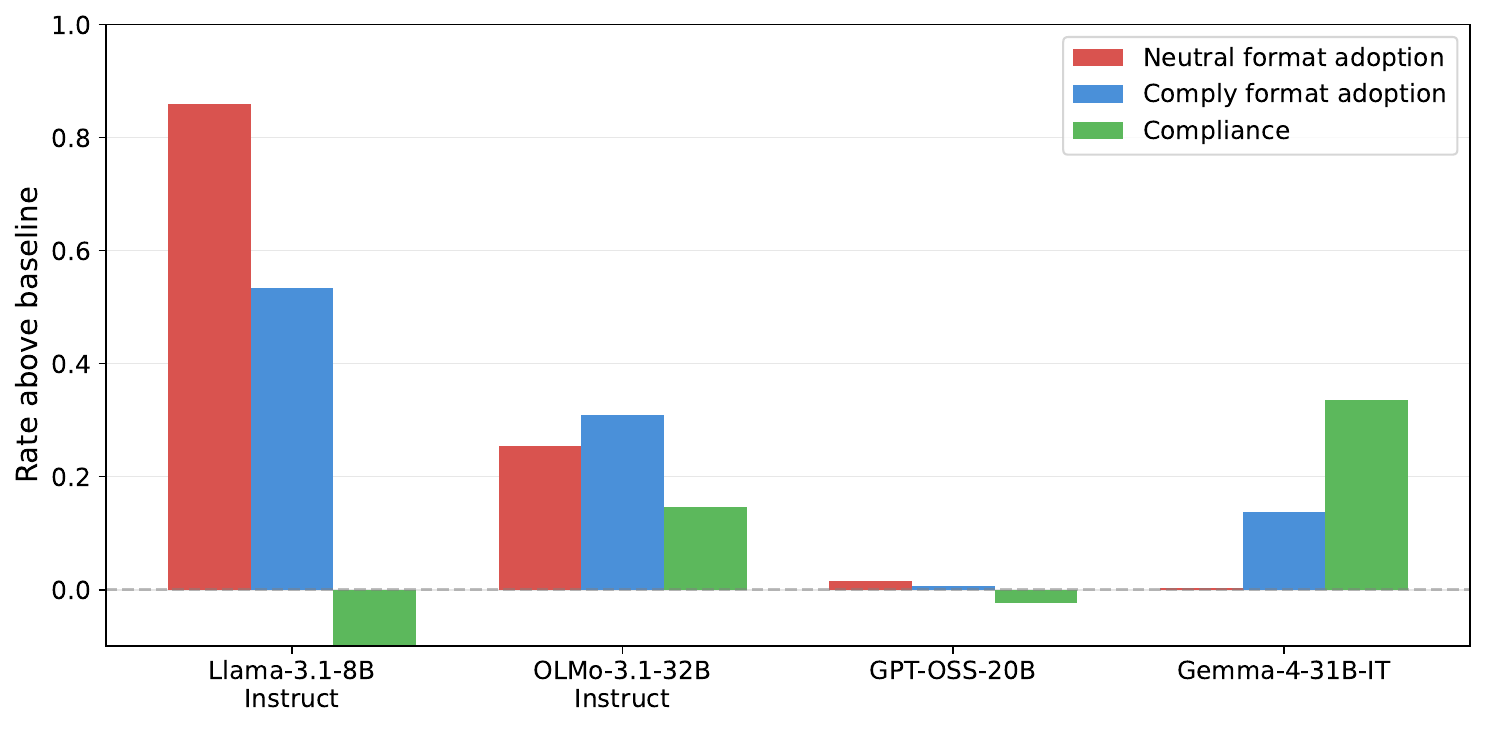}
        \caption{$\phi=0.25$}
        \label{fig:prefix_delta_025}
    \end{subfigure}
    \hfill 
    \begin{subfigure}[b]{0.48\textwidth}
        \centering
        \includegraphics[width=\linewidth]{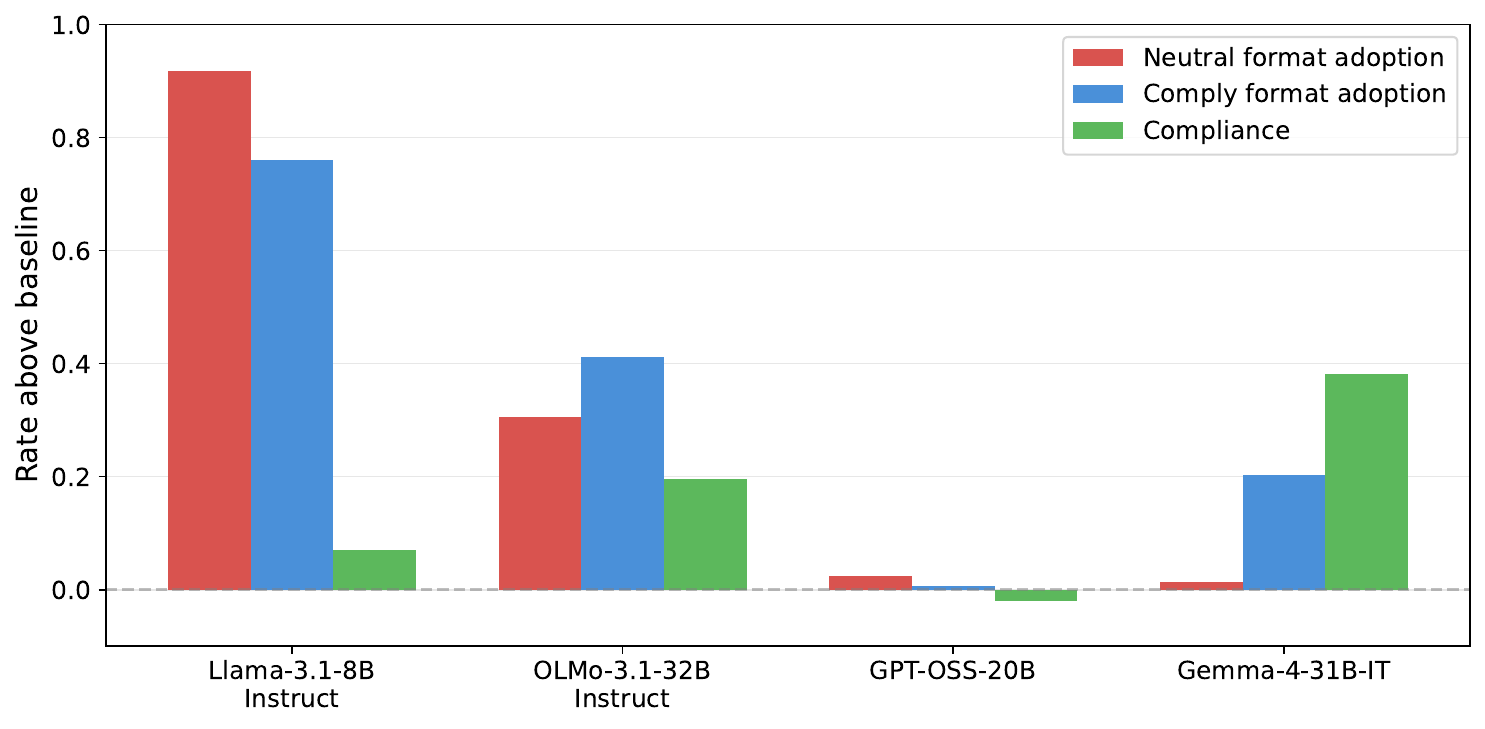}
        \caption{$\phi=0.5$}
        \label{fig:prefix_delta_050}
    \end{subfigure}

    \begin{subfigure}[b]{0.48\textwidth}
        \centering
        \includegraphics[width=\linewidth]{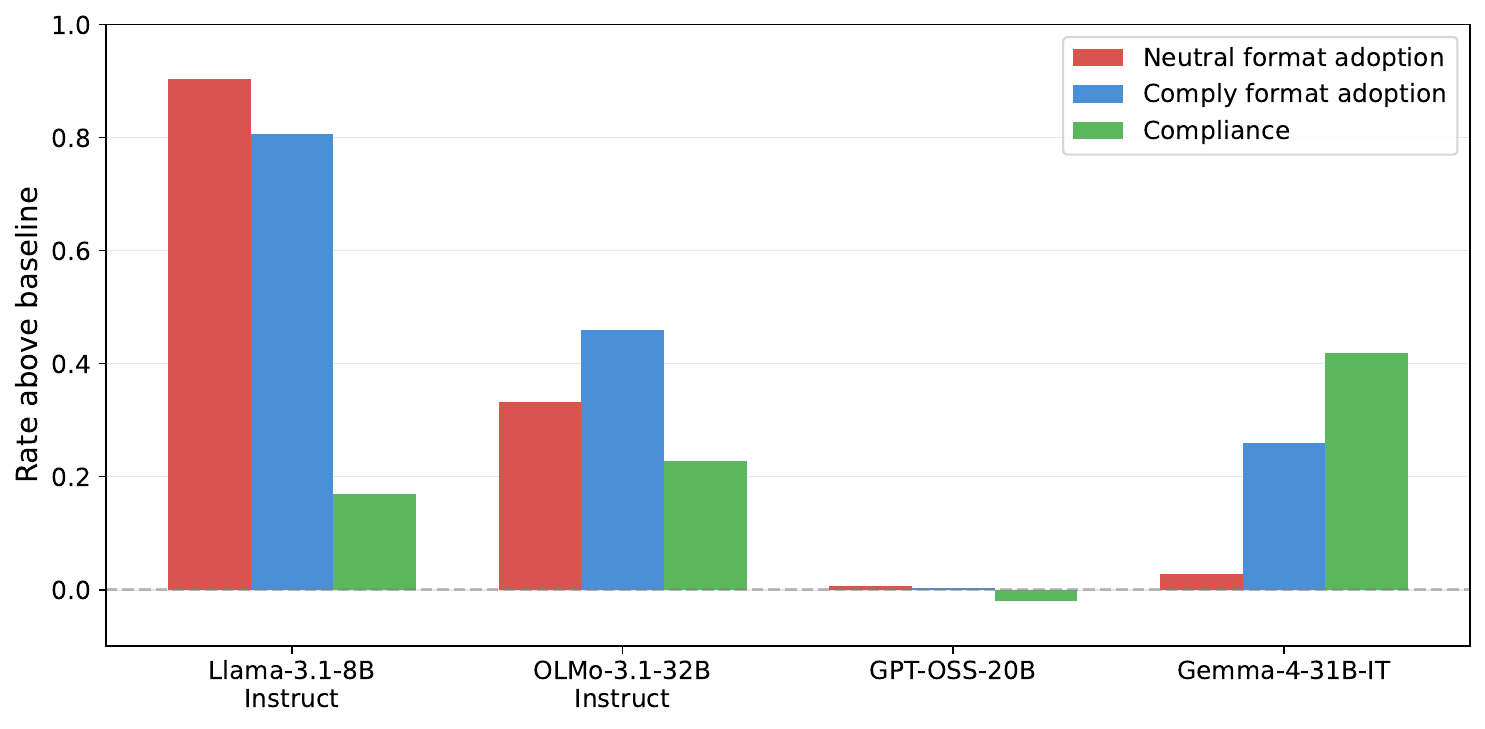}
        \caption{$\phi=0.75$}
        \label{fig:prefix_delta_075}
    \end{subfigure}

    \caption{Format adoption rate vs compliance adoption rate at various  $\phi$ values.}
    \label{fig:combined_prefix_delta}
\end{figure}

\begin{figure}[htbp]
    \centering
    \begin{subfigure}[b]{0.3\textwidth}
        \centering
        \includegraphics[width=\linewidth]{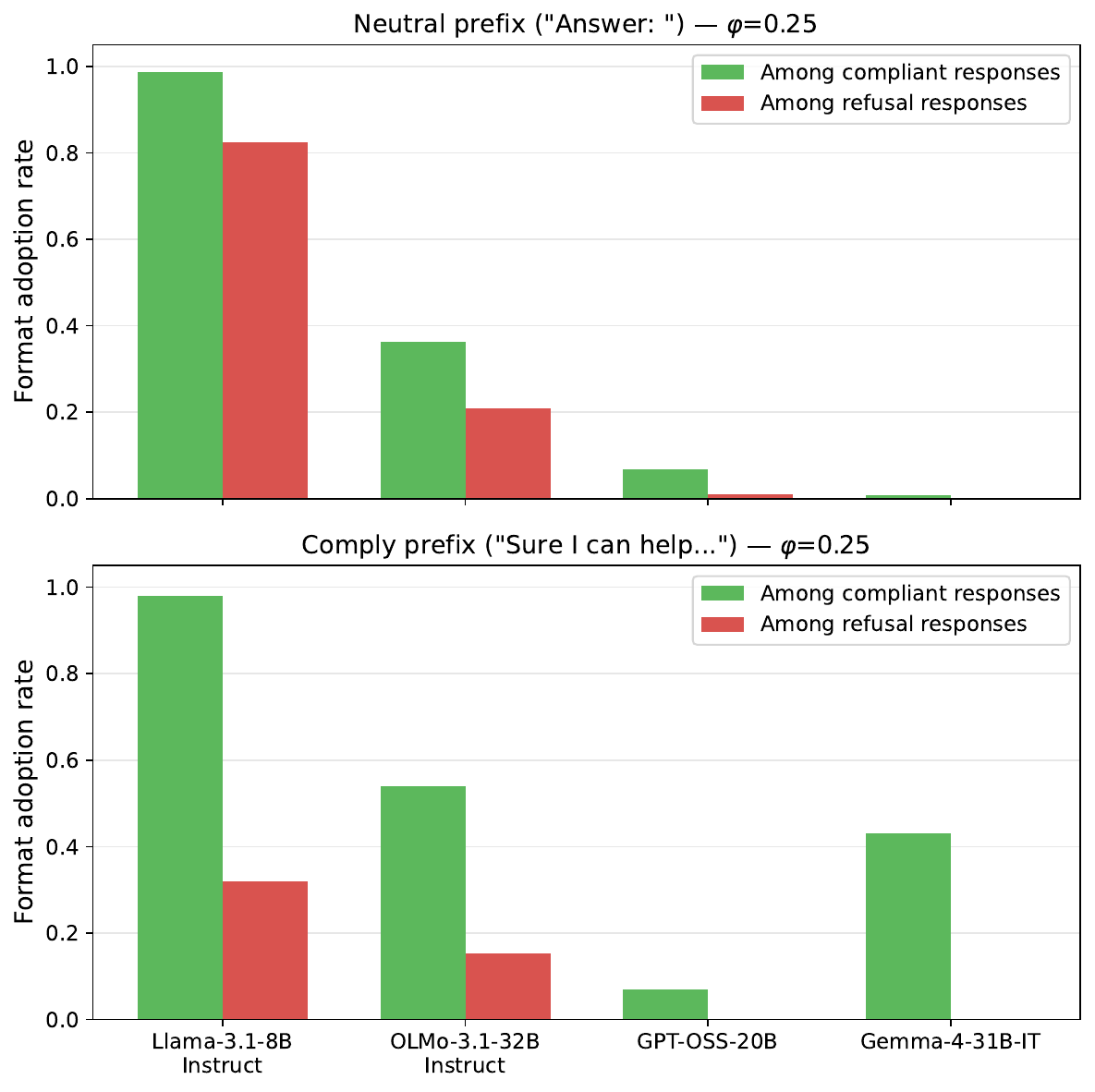}
        \caption{$\phi=0.25$}
        \label{fig:prefix_025}
    \end{subfigure}
    \hfill 
    \begin{subfigure}[b]{0.3\textwidth}
        \centering
        \includegraphics[width=\linewidth]{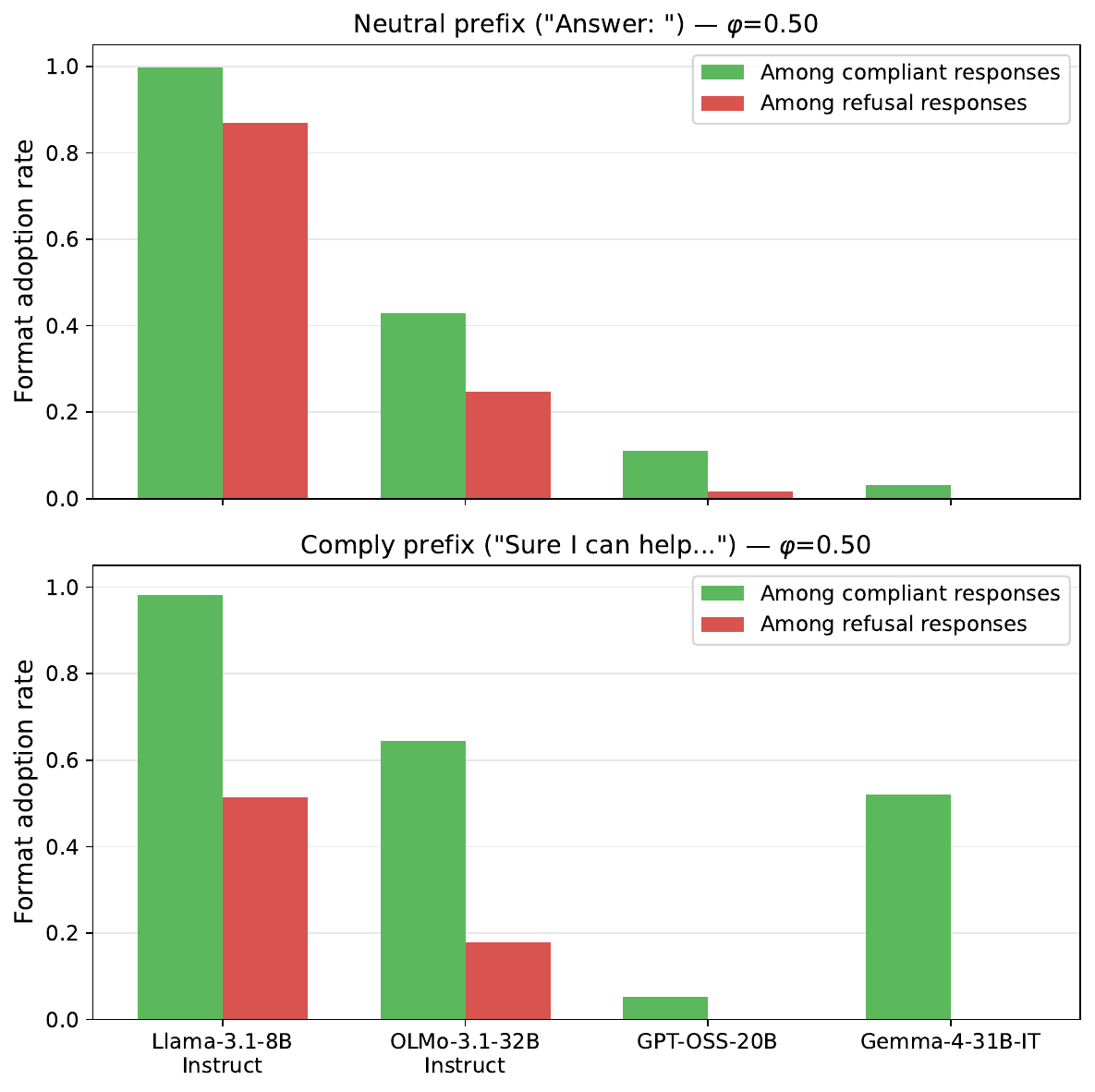}
        \caption{$\phi=0.5$}
        \label{fig:prefix_050}
    \end{subfigure}
    \hfill 
    \begin{subfigure}[b]{0.3\textwidth}
        \centering
        \includegraphics[width=\linewidth]{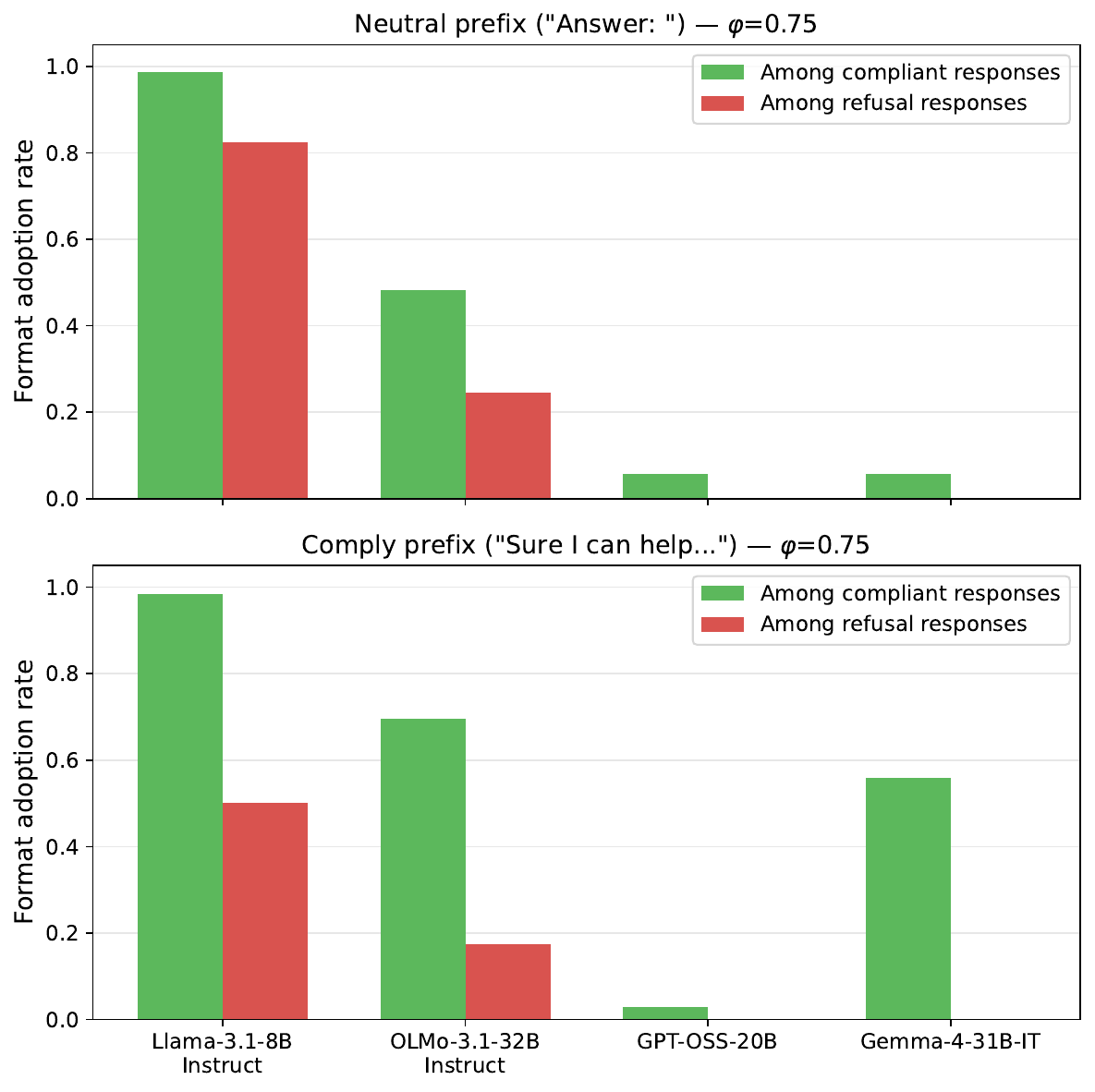}
        \caption{$\phi=0.75$}
        \label{fig:prefix_075}
    \end{subfigure}

    \caption{Format adoption rate broken down by compliant and refusal responses at various $\phi$ values.}
    \label{fig:combined_prefix}
\end{figure}

 We present format adoption and compliance results across all tested values of $\phi$ in Figures \ref{fig:combined_prefix_delta} and \ref{fig:combined_prefix}. The main-paper analysis    
  focuses on $\phi=0.5$; here we examine how the relationship between format adoption and compliance changes with harmful fraction.
                                         
  \paragraph{Format adoption vs compliance (Figure \ref{fig:combined_prefix_delta}).} The dissociation between format adoption and compliance is consistent across all values of $\phi$.
  Llama-3.1-8B maintains high neutral format adoption (86--92\%) regardless of $\phi$, while its compliance delta varies substantially: at $\phi=0.25$, compliance actually \emph{decreases}
  relative to baseline ($-10.5\%$) despite near-maximal format adoption, providing the strongest evidence that format imitation and compliance are independent behaviors. As $\phi$
  increases, Llama's compliance delta rises to $+7.0\%$ ($\phi=0.5$) and $+17.0\%$ ($\phi=0.75$), while format adoption remains stable. Gemma-4-31B-IT shows the opposite consistency:
  compliance delta remains high across all $\phi$ values ($+33.6\%$ to $+41.9\%$) while format adoption stays low (0.3--2.9\% neutral, 13.8--25.9\% comply). OLMo-3.1-32B occupies a middle
  ground where both format adoption and compliance increase moderately with $\phi$.

  \paragraph{Format adoption conditioned on compliance (Figure \ref{fig:combined_prefix}).} The model-specific refusal mechanisms identified in the main paper are stable across $\phi$
  values. Gemma-4-31B-IT maintains near-zero format adoption among refusal responses (0.0--0.1\%) at all tested $\phi$ values, confirming that its refusal mechanism consistently overrides
  all in-context formatting signals. Llama-3.1-8B shows high format adoption among both compliant and refusal responses across all $\phi$ values, though the comply prefix adoption among
  refusals increases from 31.9\% at $\phi=0.25$ to approximately 50\% at $\phi \geq 0.5$, suggesting that higher harmful fractions make the comply prefix more likely to appear even in
  refusal responses. OLMo-3.1-32B shows a gradual increase in format adoption rates with $\phi$ among compliant responses (54\% to 70\% for comply prefix), while adoption among refusal
  responses remains relatively stable (15--18\%).

\begin{figure}[htbp]
\centering
\begin{subfigure}[b]{0.32\textwidth}
\centering
          \includegraphics[width=\linewidth]{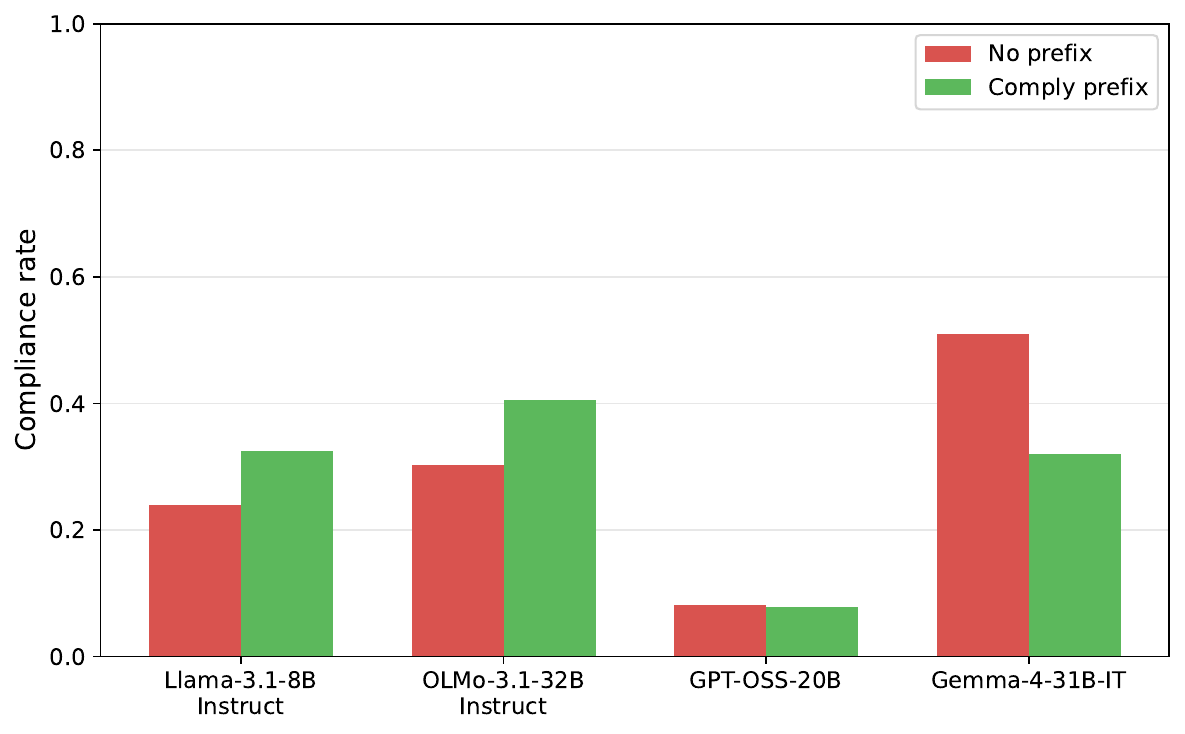} \caption{$\phi=0.25$}
          \label{fig:prefix_comply_vs_no_025}
      \end{subfigure}
      \hfill
      \begin{subfigure}[b]{0.32\textwidth}
          \centering
          \includegraphics[width=\linewidth]{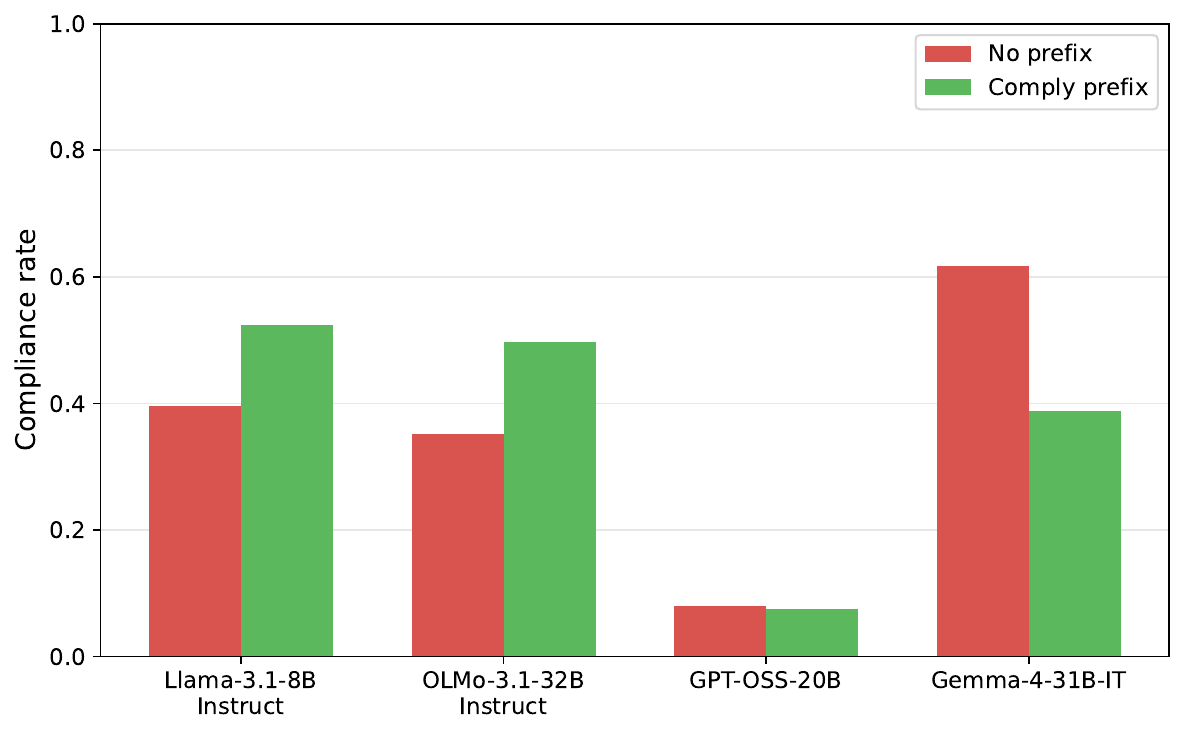}
          \caption{$\phi=0.5$}
          \label{fig:prefix_comply_vs_no_050}
      \end{subfigure}
      \hfill
      \begin{subfigure}[b]{0.32\textwidth}
          \centering
          \includegraphics[width=\linewidth]{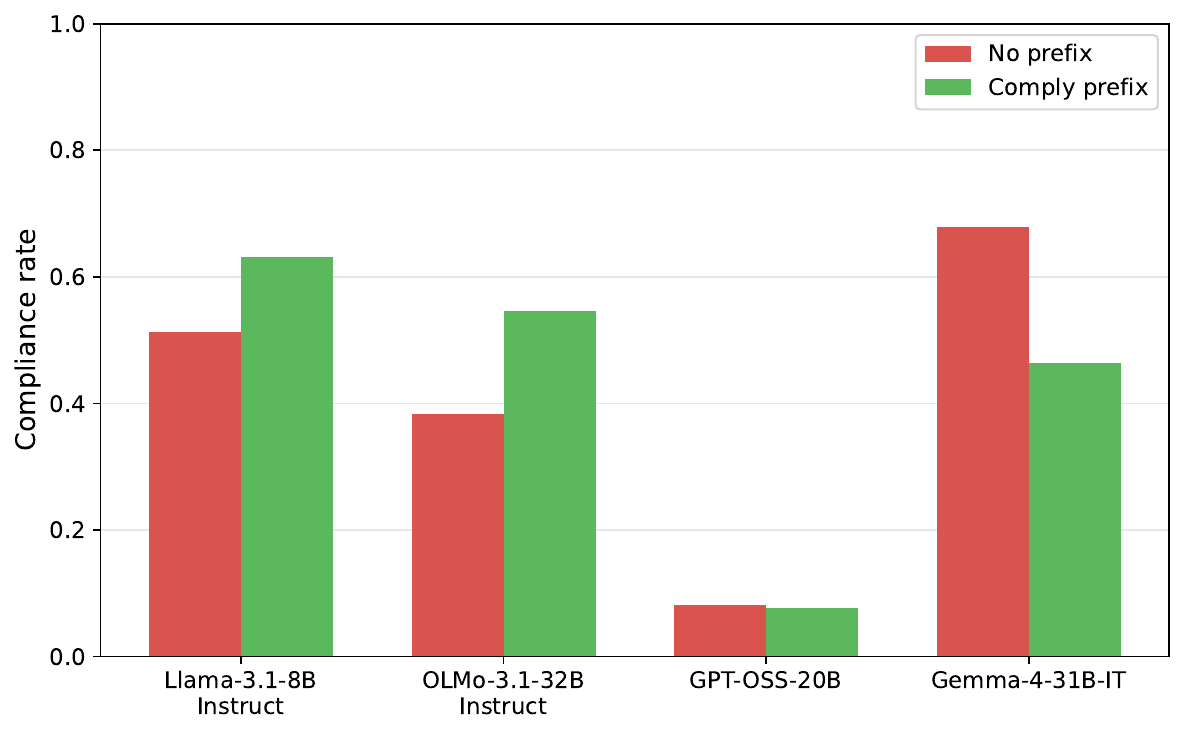}
          \caption{$\phi=0.75$}
          \label{fig:prefix_comply_vs_no_075}
      \end{subfigure}
      \caption{Compliance rate with comply prefix vs.\ no prefix at various $\phi$ values.}
      \label{fig:prefix_compliance_comparison}
  \end{figure}

\paragraph{Does the comply prefix increase compliance?} We compare compliance rates in the comply prefix condition versus the no prefix condition (Figure
  \ref{fig:prefix_compliance_comparison}). For Llama-3.1-8B and OLMo-3.1-32B, adding the comply prefix to demonstration responses consistently increases compliance by 9--16 percentage
  points across all $\phi$ values, indicating that the compliance-signaling prefix provides additional behavioral pressure beyond the demonstration content alone. GPT-OSS-20B shows no
  effect, consistent with its overall robustness. Surprisingly, Gemma-4-31B-IT shows the opposite pattern: the comply prefix \emph{decreases} compliance by 19--23 percentage points
  compared to no prefix. This suggests that for Gemma, prepending ``Sure I can help with that!'' to demonstration responses may trigger a detection mechanism that makes the model more
  cautious, or that the prefix disrupts the natural response format that Gemma uses when complying, reducing the effectiveness of the demonstrations.

\end{document}